\documentclass[11pt]{article}

\usepackage[margin=1in]{geometry}
\usepackage{amsmath,amssymb,amsfonts,bm,mathtools}
\usepackage{amsthm}
\usepackage{hyperref}
\usepackage{xcolor}
\usepackage{enumitem}
\usepackage{graphicx}
\usepackage{booktabs}
\usepackage{subcaption}
\usepackage{comment}
\usepackage{tikz}

\hypersetup{
  colorlinks=true,
  linkcolor=blue,
  citecolor=blue,
  urlcolor=blue
}

\newtheorem{remark}{Remark}

\newcommand{\R}{\mathbb{R}}

\newcommand{\cN}{{\cal N}}
\newcommand{\cG}{{\cal G}}

\graphicspath{{figs/}}

\title{Temporal Memory for Resource-Constrained Agents:\\
Continual Learning via Stochastic Compress-Add-Smooth}
\author{Michael (Misha) Chertkov\\ Graduate Inter-Disciplinary Program in Applied Mathematics,\\
\& Department of Mathematics, University Arizona, Tucson, AZ 85721}
\date{\today}

\begin{document}
\maketitle

\begin{abstract}

An agent that operates sequentially must incorporate new experience without forgetting old experience, under a fixed memory budget. We propose a framework in which memory is not a parameter vector but a stochastic process: a Bridge Diffusion on a replay interval $[0,1]$, whose terminal marginal encodes the present and whose intermediate marginals encode the past. New experience is incorporated via a three-step \emph{Compress--Add--Smooth} (CAS) recursion. We test the framework on the class of models with marginal probability densities modeled via Gaussian mixtures of fixed number of components~$K$ in $d$ dimensions; temporal complexity is controlled by a fixed number~$L$ of piecewise-linear protocol segments whose nodes store Gaussian-mixture states. The entire recursion costs $O(LKd^2)$ flops per day --- no backpropagation, no stored data, no neural networks --- making it viable for controller-light hardware.

Forgetting in this framework arises not from parameter interference but from lossy temporal compression: the re-approximation of a finer protocol by a coarser one under a fixed segment budget.  We find that the retention half-life scales linearly as $a_{1/2}\approx c\,L$ with a constant $c>1$ that depends on the dynamics but not on the mixture complexity~$K$, the dimension~$d$, or the geometry of the target family.  The constant~$c$ admits an information-theoretic interpretation analogous to the Shannon channel capacity.  The stochastic process underlying the bridge provides temporally coherent ``movie'' replay --- compressed narratives of the agent's history, demonstrated visually on an MNIST latent-space illustration.  The framework provides a fully analytical ``Ising model'' of continual learning in which the mechanism, rate, and form of forgetting can be studied with mathematical precision.
\end{abstract}

\section{Introduction}
\label{sec:introduction}

\paragraph{The agent problem.} Consider an agent --- a building controller, a robot, a sensor node --- that processes a stream of daily experiences, each represented as a probability distribution over a $d$-dimensional physical or latent state space. The agent must maintain a fixed-size memory from which it can replay past experiences to inform current decisions: warm-starting a recovery controller from last winter's occupancy pattern, recalling a previously visited room's obstacle layout, or restoring a sensor calibration profile.

The core difficulty — that a network trained sequentially on new data abruptly loses performance on previously learned tasks, a phenomenon termed {\bf catastrophic interference} \cite{mccloskey_catastrophic_1989}  or {\bf catastrophic forgetting} \cite{french_catastrophic_1999} -- has motivated a large body of work. Standard {\bf Continual Learning} (CL) methods~\cite{kirkpatrick_overcoming_2017,shin_continual_2017,schwarz_progress_2018} represent memory as neural network parameters and manage forgetting through regularisation, replay buffers -- including recent approaches that use denoising diffusion models as the replay generator \cite{gao_ddgr_2023, jodelet_class-incremental_2023, meng_diffclass_2024, kim_sddgr_2024, liang_diffusion-driven_2024, he_continual_2024, hu_replaycad_2025}, or architecture expansion~\cite{parisi_continual_2019,de_lange_continual_2022,wang_comprehensive_2024}. These approaches require gradient-based training, stored data, and compute budgets that are often unavailable on edge hardware. We propose an alternative: {\bf memory is not a parameter vector but a \emph{stochastic process} whose intermediate-time marginals encode the past}.

\paragraph{The idea.} The agent maintains a Bridge Diffusion (BD) description on a fixed replay interval $[0,1]$. The terminal marginal at $t=1$ represents the current day. Earlier days are stored as intermediate-time marginals at designated readout times $t_{m|n}\in(0,1)$. Incorporating a new day is a three-step recursion --- \emph{\bf Compress--Add--Smooth} --- carried out entirely within a chosen parameterised density class. Fixed memory is enforced by two budgets: a \emph{state budget}~$K$ (number of Gaussian-mixture components) and a \emph{temporal budget}~$L$ (number of piecewise-linear protocol segments, whose $L+1$ nodes each store a $K$-component Gaussian mixture). The total memory footprint is $O(Kd^2 L)$ floating-point numbers.

Replaying $0\to 1$ produces a compressed ``movie'' of the agent's history, realised on two levels: a smooth-in-time evolution of the marginal probability density, and smooth individual sample paths generated via the drift reconstructed from the density path (Appendix~\ref{app:p-to-s}).

\paragraph{Two stories under one mathematical umbrella.} This paper serves two audiences, unified by a common mathematical language rooted in non-equilibrium statistical mechanics, stochastic optimal control, and optimal transport theory. The mathematical backbone is a Bridge Diffusion --- a framework for controlled stochastic processes whose marginal density path is prescribed and whose drift is reconstructed from the Fokker--Planck equation. The approach is related to Schr\"odinger bridges~\cite{leonard_survey_2013,chen_stochastic_2021,de_bortoli_diffusion_2021}, flow matching~\cite{lipman_flow_2023}, stochastic interpolants \cite{albergo_building_2023} and Path-Integral Diffusion~\cite{behjoo_harmonic_2025,chertkov_adaptive_2025,chertkov_generative_2025,chertkov_mean-field_2026}, but differs in that the density path is specified directly as a piecewise-linear interpolant, rather than optimised or learned. The approach is plug-and-play: the Compress--Add--Smooth recursion works with any parameterised density family for which piecewise-linear interpolation is well defined. We illustrate it here on the simple and analytically transparent-mixture (GM) class, but the same recursion applies, in principle, to richer representations --- for instance, normalising flows or score-based models that use neural networks as function approximators.

For the \emph{\bf controls/robotics/edge-AI community}~\cite{schwarz_progress_2018,parisi_continual_2019}, the framework is a practical temporal memory for resource-constrained agents: the compress--add--smooth recursion costs $O(LKd^2)$ flops per day (matrix operations, no backpropagation), the replay query costs $O(Kd^2)$ (a single interpolation of GM parameters), and the entire pipeline runs on a microcontroller.

For the \emph{\bf continual learning community}~\cite{kirkpatrick_overcoming_2017,de_lange_continual_2022,wang_comprehensive_2024}, the GM instantiation of the framework is an analytically tractable ``Ising model'' of forgetting: a minimal, exactly solvable system in which the mechanism (temporal compression), rate (controlled by  $L$), and form (two-regime curve, confusion-dominated) of forgetting can be studied with mathematical precision --- questions that are not feasible to answer in neural-network-based and dynamics-absent CL. In the neuroscience-inspired sleep replay literature~\cite{gonzalez_can_2020,golden_sleep_2022,tadros_sleep-like_2022,golden_interleaved_2025,vins_optimal_2025}, off-line replay is shown to prevent catastrophic forgetting by pushing synaptic weights toward joint solution manifolds; our SDE-based replay (Section~\ref{sec:stochastic_replay}) is structurally analogous.

\paragraph{Summary of results.}
We report experiments for single-Gaussian ($K{=}1$), Gaussian-Mixture (GM) ($K$ up to~8), and MNIST \cite{lecun_mnist_1998} latent-space GM daily distributions over $n{=}100$ days, with the following principal findings:
\begin{enumerate}[nosep]
\item \emph{Two-regime forgetting curve.}  The normalized forgetting $\bar F(a)$ exhibits a low-error plateau for recent memories, followed by a steep sigmoid transition.  The retention half-life $a_{1/2}$ --- the age at which $\bar F$ crosses~$0.5$ --- is the natural summary statistic (Section~\ref{sec:phase1}).

\item \emph{Linear scaling: $a_{1/2}\approx c\,L$.}  The half-life scales linearly with the segment budget, from $a_{1/2}=14$ at $L=5$ to $a_{1/2}=74$ at $L=30$, with $c\approx 2.4$ for the default geometry. Since $c>1$, the CAS scheme outperforms a na\"ive First-In-First-Out (FIFO) buffer (which gives $a_{1/2}=L$) by a factor of~${\sim}2.4\times$.  We argue that~$c$ admits an information-theoretic interpretation as a channel capacity (Section~\ref{sec:capacity}).  This linear scaling is confirmed across all experimental settings (Sections~\ref{sec:phase2}--\ref{sec:scaling}).

\item \emph{Independence of~$K$, $d$, and geometry.}  The half-life is essentially independent of the mixture complexity~$K$ (tested for $K\in\{1,2,3,5,8\}$), the ambient dimension~$d$ (tested for $d$ up to~30), the crowding geometry, and even topological curriculum changes (split-merge events).  Only drift speed has a measurable effect, modulating~$c$ from ${\sim}2.0$ (fast drift) to ${\sim}3.6$ (slow drift).

\item \emph{Confusion, not destruction.}  Old memories collapse toward recent eras ($\bar F > 1$) rather than reverting to the prior ($\bar F \to 1$).  This ``confusion'' regime is the dominant failure mode.

\item \emph{Adaptive forgetting channel.}  The decomposed metric identifies the active information channel: mean-dominated (${\sim}85\%$) when component means drift (synthetic), covariance-dominated when only weights vary (MNIST).  Weight error is negligible for equal-weight mixtures.

\item \emph{Movie replay.}  The stochastic process reconstructed from the density path produces temporally coherent replay trajectories --- compressed ``movies'' of the agent's history.  On MNIST \cite{lecun_mnist_1998}, the protocol grid decoded frame-by-frame produces a visual temporal narrative in which digit identities are preserved (Section~\ref{sec:mnist_latent}).
\end{enumerate}

\paragraph{Paper outline.} Section~\ref{sec:framework} introduces the CAS recursion. Section~\ref{sec:forgetting_distinction} identifies forgetting-by-compression. Section~\ref{sec:metrics} defines the forgetting metrics. Sections~\ref{sec:phase1}--\ref{sec:scaling} report experiments. Section~\ref{sec:discussion} discusses the capacity law and stochastic replay.  Section~\ref{sec:mnist_latent} presents the MNIST illustration. Section~\ref{sec:conclusions} concludes. Appendix~\ref{app:p-to-s} derives the drift from the density path; Appendix~\ref{app:design} describes the software architecture.

\section{The Compress--Add--Smooth Framework}
\label{sec:framework}

The agent maintains a Bridge Diffusion (BD) process on the fixed replay interval $[0,1]$ whose terminal marginal at $t=1$ represents the current day, and whose intermediate-time marginals encode the past.  Incorporating a new day is a three-step recursion --- compress, add, smooth --- carried out entirely within a chosen parameterised density class.  The approach is generic: it applies to any density family for which piecewise-linear interpolation is well defined.  We illustrate it in this paper on the Gaussian-Mixture (GM) class, where all operations reduce to linear algebra on the mixture parameters. The protocol grid is kept \emph{uniform} at all times: after every daily update the node times are $\{0, 1/L, 2/L, \ldots, 1\}$.  We achieve this by {\bf compressing} at every time step the domain from $[0,1]$ to $[0, L/(L+1)]$, then fitting the new day's experience in the newly {\bf added} interval $[L/(L+1),1]$, and then {\bf smoothing} the resulting $L+1$ segments back to $L$ segments. 

\begin{figure}[h!]
\centering
\includegraphics[width=\textwidth]{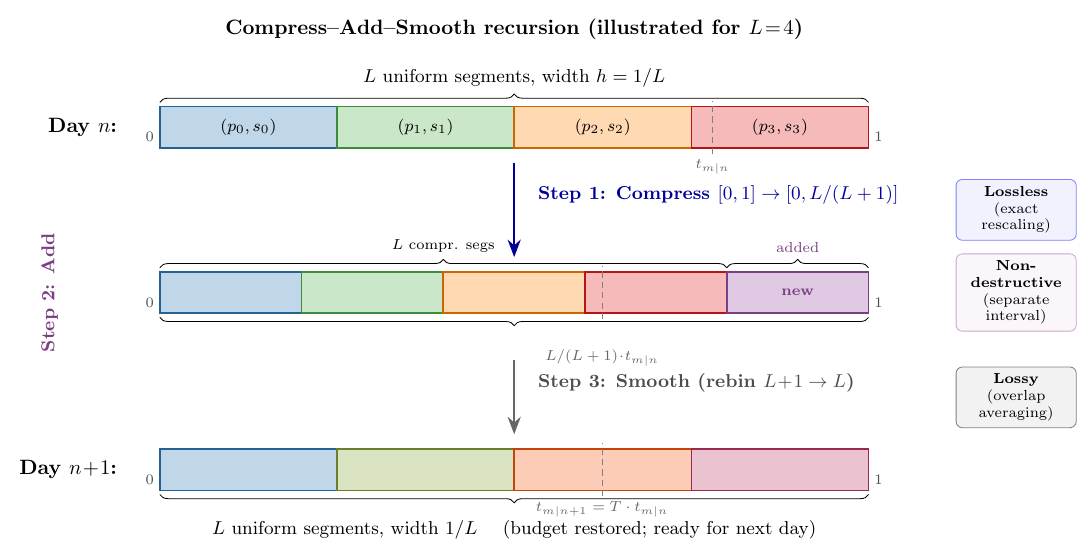}
\caption{One iteration of the compress--add--smooth recursion, illustrated for $L\!=\!4$ segments. \textbf{Top:} the protocol at day~$n$ consists of $L$ uniform segments on $[0,1]$. \textbf{Middle:} compression rescales the protocol to $[0,L/(L+1)]$; the new day is appended on $[L/(L+1)]$, producing $L\!+\!1$ uniform segments. \textbf{Bottom:} rebinning averages the $L\!+\!1$ segments back onto the $L$-segment grid. The right-hand labels indicate the information-theoretic role of each step: only smoothing is lossy. Dashed lines track a past-day readout time $t_{m|n}$, which contracts by factor $L/(L+1)$ every day.}
\label{fig:scheme_b}
\end{figure}

Fig.~\ref{fig:scheme_b} illustrates the recursion for $L=4$.

\subsection{Memory representation}
\label{sec:memory_rep}

At day~$n$, the agent's memory consists of three objects:
\begin{enumerate}[label=(\roman*),nosep]
\item a \emph{prior distribution} $q^{(0)}\in\cG_K$, where $\cG_K$ is the class of
      $K$-component Gaussian mixtures;

\item a \emph{protocol grid}: $L+1$ Gaussian-mixture states $\{G_j^{(n)}\}_{j=0}^{L}$, one at each node time $t_j = j/L$.  Each $G_j^{(n)}\in\cG_K$ is specified by weights $\pi_k^{(j)}$, means $m_k^{(j)}\in\R^d$, and covariances $\Sigma_k^{(j)}\in\R^{d\times d}$, $k=1,\ldots,K$.  Between adjacent nodes the density is defined by piecewise-linear interpolation of the GM parameters (see below);

\item a readout-time dictionary $\{t_{m|n}\}_{m=1}^{n}$, mapping each past day~$m$ to a query time in $(0,1]$.
\end{enumerate}
The total memory cost is $O(LKd^2)$ for the protocol grid ($L+1$ nodes, each storing $K$ means of size $d$ and $K$ covariance matrices of size $d^2$) plus $O(Kd^2)$ for the prior.

\paragraph{Piecewise-linear interpolation.} For any $t\in[t_j, t_{j+1}]$, with $\alpha = (t - t_j)/(t_{j+1} - t_j)$, the marginal density is the Gaussian mixture with linearly interpolated parameters:
\begin{equation}
  \pi_k(t) = (1-\alpha)\,\pi_k^{(j)} + \alpha\,\pi_k^{(j+1)},
  \quad
  m_k(t) = (1-\alpha)\,m_k^{(j)} + \alpha\,m_k^{(j+1)},
  \quad
  \Sigma_k(t) = (1-\alpha)\,\Sigma_k^{(j)} + \alpha\,\Sigma_k^{(j+1)}.
  \label{eq:pw_interp}
\end{equation}
This interpolation preserves the GM structure: for every $t$, the marginal is a valid $K$-component Gaussian mixture (weights sum to~1, covariances are positive definite by convexity). The corresponding SDE drift, needed only when sample paths are required, is reconstructed from the density path via Appendix~\ref{app:p-to-s}.

\subsection{Initialisation (day 1)}
\label{sec:init}

The initial (day 1) protocol is set up by linearly interpolating from the prior distribution $q^{(0)}$ at $t=0$ to the first day's target $q^{(1)}$ at $t=1$:
\begin{equation}\label{eq:p-first-day}
p_t^{(1)}=(1-t) q^{(0)} + t q^{(1)},
\end{equation}
where the linear combination acts on the GM parameters as in~\eqref{eq:pw_interp}.
The $L+1$ initial node states are obtained by evaluating this interpolant at the grid times $t_j = j/L$:
\begin{equation}
  G_j^{(1)} = \bigl(1 - j/L\bigr)\, q^{(0)} + \bigl(j/L\bigr)\, q^{(1)},
  \qquad j = 0,\ldots,L.
\end{equation}
(A nonlinear interpolant can also be used.) The drift corresponding to the density path~\eqref{eq:p-first-day} is reconstructed via Appendix~\ref{app:p-to-s}.

\subsection{Step 1: exact compression}
\label{sec:compress}

The old protocol, defined on $L+1$ nodes at times $\{0, 1/L, \ldots, 1\}$, is mapped exactly to the subinterval $[0,L/(L+1)]$ by relabelling the node times:
\begin{equation}
  G_j^{(n+1,\text{cmp})} = G_j^{(n)},
  \qquad
  \text{at time } t_j^{\text{cmp}} = \frac{j}{L}\cdot\frac{L}{L+1} = \frac{j}{L+1},
  \qquad j = 0,\ldots,L.
\end{equation}
The GM states at each node are unchanged; only the time labels are rescaled.  This is an exact, lossless operation: the compressed protocol defines the same density path, played at $L/(L+1)$ speed.

\subsection{Step 2: addition}
\label{sec:add}

A single new node is appended at $t=1$ with state $q^{(n+1)}$ (the new day's target distribution).  The compressed grid already has a node at $t = L/(L+1)$ with state $G_L^{(n)}$ (the previous day's terminal marginal). Between these two nodes the density is again defined by linear interpolation:
\begin{equation}\label{eq:p-add}
p_t^{(n+1)} = \frac{1-t}{1-L/(L+1)}\, G_L^{(n)} + \frac{t - L/(L+1)}{1-L/(L+1)}\, q^{(n+1)},
\qquad t\in\bigl[L/(L+1),\,1\bigr].
\end{equation}
After addition, the augmented protocol has $L+2$ nodes at the uniform grid $\{0, \tfrac{1}{L+1}, \tfrac{2}{L+1}, \ldots, \tfrac{L}{L+1}, 1\}$, constituting $L+1$ segments of width $1/(L+1)$.

\subsection{Step 3: smoothing by uniform-grid rebinning}
\label{sec:smooth}

The augmented protocol has $L+2$ nodes (constituting $L+1$ segments), but the budget allows only~$L$ segments ($L+1$ nodes). We restore the budget by \emph{rebinning}: evaluating the augmented piecewise-linear density interpolant at the target grid and storing the resulting GM states as the new nodes.

Concretely, the augmented grid has nodes at $t_k^{\rm aug} = k/(L+1)$, $k=0,\ldots,L+1$, and the target grid has nodes at $t_j^{\rm new} = j/L$, $j=0,\ldots,L$.  For each target node, we evaluate the augmented interpolant:
\begin{equation}
  G_j^{\rm new} = p_{t_j^{\rm new}}^{\rm aug},
  \qquad j = 0,\ldots,L.
  \label{eq:rebin_eval}
\end{equation}
Since $t_j^{\rm new}$ falls inside some augmented segment $[t_k^{\rm aug}, t_{k+1}^{\rm aug}]$, the evaluation is a linear interpolation between two adjacent augmented nodes:
\begin{equation}
  G_j^{\rm new} = (1-\alpha_{jk})\, G_k^{\rm aug} + \alpha_{jk}\, G_{k+1}^{\rm aug},
  \qquad
  \alpha_{jk} = \frac{t_j^{\rm new} - t_k^{\rm aug}}{t_{k+1}^{\rm aug} - t_k^{\rm aug}},
  \label{eq:rebin_interp}
\end{equation}
where $k$ is the unique index such that $t_k^{\rm aug} \le t_j^{\rm new} < t_{k+1}^{\rm aug}$.  Since the interpolation acts componentwise on the GM parameters $(\pi, m, \Sigma)$, the result is a valid $K$-component Gaussian mixture at every node (weights are convex combinations summing to~1; covariances remain positive definite by convexity of the PSD cone).

Equivalently, the operation can be written as a matrix--vector product using a sparse \emph{rebinning matrix} $W \in \R^{(L+1) \times (L+2)}$ whose rows encode the interpolation weights~\eqref{eq:rebin_interp}.  Each row of~$W$ has at most two nonzero entries and sums to~1.  The matrix depends only on~$L$ and is precomputed once.

The entire smoothing step requires $O(LKd^2)$ operations: for each of $L+1$ target nodes, interpolate the $K \times (d^2 + d + 1)$ GM parameters.  No optimiser, no merge-pair selection, and no policy choice is needed.

\subsection{Readout-time evolution}
\label{sec:readout}

Readout times are updated only in the compression step:
\begin{equation}
t_{m|n+1} = \frac{L}{L+1}\cdot t_{m|n}\qquad \text{for all } m \le n, \qquad t_{n+1|n+1} = 1.
\label{eq:readout_update}
\end{equation}
The smoothing step does not move readout times; it changes the node states of the protocol grid, so the marginal \emph{at} the readout time changes --- that is the forgetting mechanism. After $n-m$ days, the readout time of day~$m$ is
\begin{equation}
t_{m|n} = \Bigl(\frac{L}{L+1}\Bigr)^{\!n-m},
\label{eq:readout_geometric}
\end{equation}
which decays geometrically toward~0 with age. For $L=10$, the readout time of a 20-day-old memory is $(L/(L+1))^{20} \approx 0.12$, placing it in the leftmost 12\% of the replay interval.

\subsection{Computational cost}
\label{sec:cost}

\paragraph{Per-day update:} $O(LKd^2)$. Compression relabels $L+1$ node times ($O(1)$ per node; the GM states are unchanged). Addition appends one new node and evaluates one interpolation ($O(Kd^2)$). Smoothing evaluates the augmented interpolant at $L+1$ target nodes ($O(Kd^2)$ per node), giving $O(LKd^2)$ total. No backpropagation, no sampling, no optimiser.

\paragraph{Per-replay query:} $O(Kd^2)$. The replay marginal at readout time $t_{m|n}$ is obtained by evaluating the piecewise-linear interpolant~\eqref{eq:pw_interp}: locate the enclosing segment ($O(\log L)$ or $O(1)$ with a uniform grid), then interpolate $K$ means ($Kd$ operations) and $K$ covariance matrices ($Kd^2$ operations).

\paragraph{Memory footprint:} For $d=8$, $K=3$, $L=20$: the protocol occupies $21 \times 3 \times (64+8+1) = 4599$ floats $\approx$ 37\,kB in double precision, plus $\sim$1.8\,kB for the prior. No stored data, no replay buffer.

\section{Forgetting-by-compression}
\label{sec:forgetting_distinction}

In standard continual learning, forgetting arises from \emph{parameter interference} \cite{mccloskey_catastrophic_1989,french_catastrophic_1999}: gradient updates on new data overwrite representations needed for old tasks. In our framework, the three steps have distinct information-theoretic roles:
\begin{itemize}[nosep]
\item \emph{Compression} is \textbf{lossless} --- it is an exact time-rescaling that preserves the marginal flow.

\item \emph{Addition} is \textbf{non-destructive} --- the new day occupies a separate interval $[L/(L+1),1]$ and does not modify the old protocol on $[0,L/(L+1)]$.

\item \emph{Smoothing} is \textbf{lossy} --- rebinning replaces a finer grid by a coarser one, erasing sub-grid temporal detail.
\end{itemize}
Forgetting is therefore \emph{localised in a single identifiable step}: the re-approximation of an $(L\!+\!1)$-segment protocol by an $L$-segment protocol via interpolant evaluation on a coarser grid. The temporal resolution available for old memories shrinks geometrically with age through the readout-time decay~\eqref{eq:readout_geometric}, making forgetting a consequence of \emph{temporal coarse-graining} rather than parametric interference.

\begin{remark}[Temporal blurring of node states]
\label{rem:gradient}
Each rebinning cycle replaces node states by convex combinations of their neighbors, progressively smoothing the spatial variation along the protocol. Older (leftward) nodes have undergone more rebinning cycles and their GM parameters are therefore more blurred --- component means are pulled toward a common average, covariances are inflated, and weight contrasts are reduced. This cumulative blurring is the microscopic mechanism behind the macroscopic forgetting curve. It can serve as a diagnostic: when leftward nodes become nearly indistinguishable, the memory is close to saturation.
\end{remark}

\section{Forgetting metrics}
\label{sec:metrics}

We use moment-based metrics as the primary forgetting diagnostics throughout this paper.  They are cheap to evaluate, analytically transparent, and sufficient for the GM class.  For richer density families (e.g.\ neural parameterisations), distributional metrics such as KL divergence or Wasserstein-2 distance would be natural alternatives; we leave their systematic study to future work.

\subsection{Raw moment mismatch}
\label{sec:metric_raw}

The replay distribution of past day~$m$ at current day~$n$ is $\widehat p^{(m|n)} = p_{t_{m|n}}^{(n)}$, the marginal of the current protocol evaluated at the readout time via~\eqref{eq:pw_interp}. The raw forgetting metric is
\begin{equation}
F_{m\to n} \;=\; \|\mu_{\rm replay} - \mu_{\rm orig}\|^2 + \|\Sigma_{\rm replay} - \Sigma_{\rm orig}\|_F^2,
\label{eq:F_raw}
\end{equation}
where $(\mu,\Sigma)$ are the overall mean and covariance of the Gaussian mixture, computed analytically from the GM parameters.

Rather than studying the full $(m,n)$ matrix, we work primarily with the \emph{age variable} $a=n-m$ and define the age-dependent forgetting curve
\begin{equation}
\bar F(a) \;=\; \bigl\langle \bar F_{m\to n} \bigr\rangle_{n-m=a}
\label{eq:F_age}
\end{equation}
as the average over all pairs with $n-m=a$.

\subsection{Normalised metric}
\label{sec:metric_norm}

To compare across $K$, $d$, and geometric scale, we normalise by the \emph{amnesia baseline}:
\begin{equation}
F_{\rm amnesia}(m) = \|\mu_0 - \mu_{\rm orig}^{(m)}\|^2 + \|\Sigma_0 - \Sigma_{\rm orig}^{(m)}\|_F^2,
\label{eq:F_amnesia}
\end{equation}
where $(\mu_0,\Sigma_0)$ are the moments of the starting distribution (for a deterministic start: $\mu_0=x_0$, $\Sigma_0=0$). The normalised forgetting is
\begin{equation}
\bar F_{m\to n} = \frac{F_{m\to n}}{F_{\rm amnesia}(m)}\;\in\;[0,\infty).
\label{eq:F_norm}
\end{equation}
Here $\bar F=0$ is perfect recall, $\bar F=1$ is total amnesia, and $\bar F > 1$ indicates \emph{confusion} --- the replay is actively worse than having no memory at all. The retention half-life is $a_{1/2} = \min\{a : \bar F(a) \ge \theta\}$ with threshold $\theta = 0.5$.

\begin{remark}[Confusion: $\bar F > 1$]
\label{rem:confusion}
When $\bar F > 1$, old memories have been pulled toward the current day's location rather than decaying toward the prior. We call this regime \emph{confusion} to distinguish it from \emph{destruction} ($\bar F \to 1$, reversion to the uninformed prior).
\end{remark}

\subsection{Decomposed metric for Gaussian mixtures}
\label{sec:metric_decomp}

For $K>1$, we decompose forgetting into per-component contributions after Hungarian matching:
\begin{align}
F &= F_{\rm mean} + F_{\rm cov} + F_{\rm weight}, \notag \\
F_{\rm mean} &= \textstyle\sum_k \bar w_k \|\Delta m_k\|^2,\quad
F_{\rm cov} = \textstyle\sum_k \bar w_k \|\Delta\Sigma_k\|_F^2,\quad
F_{\rm weight} = \|\Delta\pi\|^2,
\label{eq:F_decomp}
\end{align}
where $\bar w_k = \max(\pi_k^{\rm replay}, \pi_k^{\rm orig})$ and matching is by pairwise mean distance.

\section{Experiments: single-Gaussian ($K=1$, $d=2$)}
\label{sec:phase1}

\subsection{Setup}

We consider a stream of $n=100$ daily Gaussian targets
\[
p^{(m)} = \mathcal N\!\bigl(\mu^{(m)}, \Sigma\bigr), \qquad m=1,\dots,n,
\]
in dimension $d=2$, with fixed covariance
\[
\Sigma = 0.5\,I.
\]
Unless stated otherwise, the daily means follow a circular drift of radius $R=2$,
\[
\mu^{(m)} = R\bigl(\cos(2\pi m/P),\, \sin(2\pi m/P)\bigr),
\qquad P=50,
\]
so that over the $100$-day horizon the mean completes two full revolutions. The prior is $q^{(0)} = \cN(0, I)$.

The default segment budget is
\[
L=10.
\]

The circular drift is a deliberately nontrivial geometry. It is simple enough to visualize, but unlike a monotone linear drift it periodically revisits earlier spatial locations. This makes it possible to separate two effects: genuine temporal forgetting and geometric aliasing caused by revisiting the same region of state space at different times. To assess the role of geometry, we also compare against a linear-drift experiment in which the daily means move along a line at comparable local speed.

\subsection{Default behavior: age curve, heatmap, and replay geometry}

\begin{figure}[h!]
\centering
\includegraphics[width=0.95\linewidth]{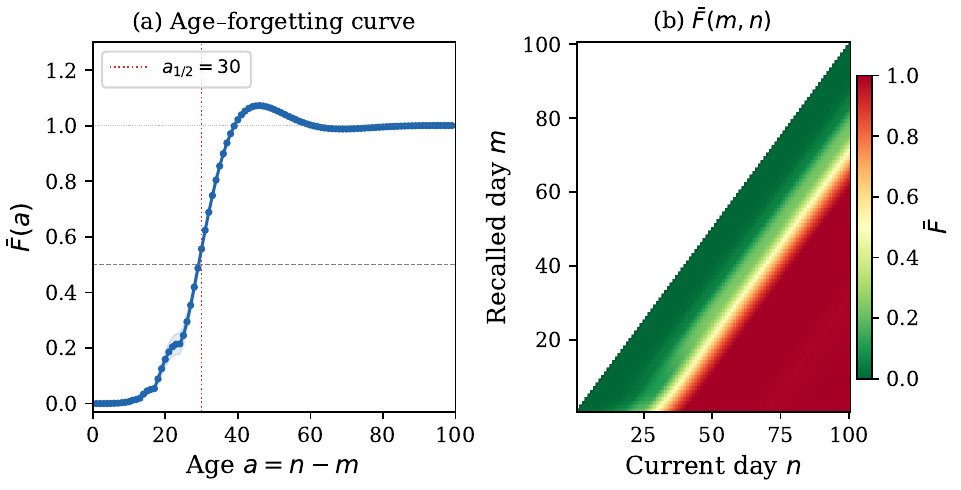}
\caption{Single-Gaussian experiment (\(K=1\), \(d=2\), \(L=10\)) under the default circular-drift setting. (a) Age-averaged normalized forgetting \(\bar F(a)\), showing retention half-life \(a_{1/2}=30\). The curve exhibits a low-error plateau for ages~$0$--$15$, followed by a steep sigmoid transition crossing $\bar F = 0.5$ at age~30, a slight overshoot to $\bar F \approx 1.08$ around age~50 (the confusion regime), and eventual saturation near $\bar F = 1.0$. The curve is weakly non-monotone due to the periodic geometry of the circular drift, which causes geometric recurrence at multiples of the half-period. (b) Full forgetting matrix \(\bar F(m,n)\) as a function of recalled day \(m\) and current day \(n\). The dominant trend is age-controlled forgetting (iso-forgetting contours run parallel to the diagonal), modulated by the periodic geometry of the underlying drift.
}
\label{fig:k1_age_heatmap}
\end{figure}

Fig.~\ref{fig:k1_age_heatmap} shows the basic forgetting diagnostics for the default parameters.  Panel~(a) reveals a characteristic two-regime structure: recent memories ($a \lesssim 15$) are recalled with near-zero error, while older memories undergo a rapid sigmoid-like degradation.  The half-life $a_{1/2} = 30$ means that, with $L=10$ segments, the agent retains useful recall of the past ${\sim}30$~days.  The slight overshoot $\bar F > 1$ in the age range $40$--$60$ confirms the confusion phenomenon (Remark~\ref{rem:confusion}): old replayed means are pulled toward the current day's location rather than decaying to the prior.  Panel~(b) shows that the dominant structure of the forgetting matrix is age-controlled, with periodic modulation visible as faint stripes at multiples of the half-period $P/2 = 25$.

\begin{figure}[h!]
\centering
\includegraphics[width=0.95\linewidth]{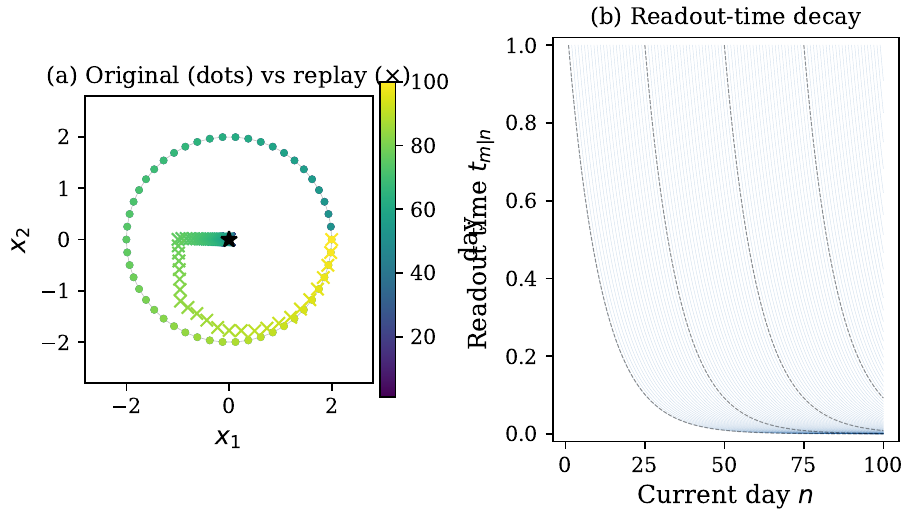}
\caption{Default single-Gaussian circular-drift experiment (\(L=10\)). (a) Original daily means (dots, coloured by day) and replayed means at the final day (crosses). The black star marks the prior mean (the origin), which serves as the protocol's long-time attractor. Confusion is visible as the systematic inward displacement of crosses from the circle toward the star: recent memories (warm colours) are replayed near their true locations on the circle, while older memories (cool colours) are pulled progressively toward the prior mean.  This convergence of replay means toward the star --- rather than remaining on the circle --- is the geometric signature of confusion. (b) Readout times \(t_{m|n}\) versus current day \(n\), showing the geometric decay~\eqref{eq:readout_geometric}.
}
\label{fig:k1_traj_readout}
\end{figure}

Fig.~\ref{fig:k1_traj_readout}(a) visualizes the replayed means at the final day $n=100$. Recent memories are replayed close to their true locations on the right-lower arc of the circle, whereas older memories are displaced toward a compressed cluster near the origin.  This spatial collapse is the geometric signature of confusion: old replayed means are attracted toward the time-weighted average of the protocol, which is dominated by recent days.

\begin{figure}[h!]
\centering
\includegraphics[width=0.82\linewidth]{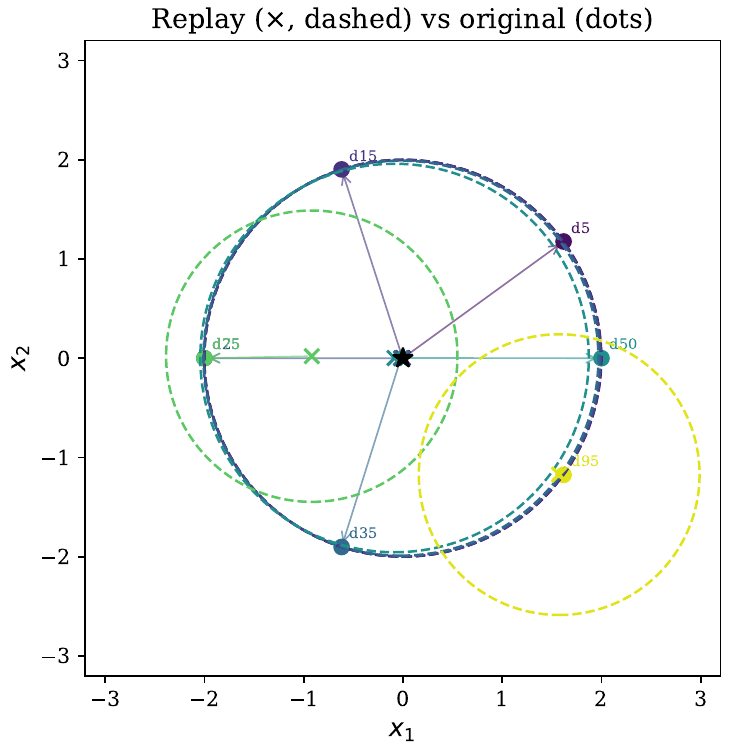}
\caption{Selected replay ellipses in the default single-Gaussian circular-drift experiment (\(L=10\)). For each displayed day, the original target mean is shown by a dot, while the replayed mean is shown by a cross with a dashed ellipse representing the replay covariance.  Recent memories (e.g.\ day~95) are replayed with small displacement and compact ellipses.  Intermediate-age memories (e.g.\ day~35) show large displacement toward the origin and inflated covariances.  Very old memories (e.g.\ day~25) collapse nearly to the origin with very large ellipses.  Geometric aliasing is visible for day~5, whose true location on the circle lies close to day~95 (they differ by two full periods), producing an apparently accurate replay that is coincidental rather than genuine recall.
}
\label{fig:k1_confusion}
\end{figure}

Fig.~\ref{fig:k1_confusion} makes the confusion mechanism visible: as age increases, the replayed mean migrates inward from the true location on the circle toward the origin (the time-averaged protocol centre), while the replayed covariance inflates dramatically.  The arrows connecting original to replayed positions show that the displacement is systematically directed toward the protocol interior, not random.

Fig.~\ref{fig:k1_traj_readout}(b) shows the readout times $t_{m|n}$ versus current day $n$. They decay geometrically as $(L/(L+1))^{n-m}$, with the theoretical curves (dashed) overlaid for reference.  The actual and theoretical curves coincide exactly, confirming the readout-time evolution~\eqref{eq:readout_geometric}.  For $L=10$, a 30-day-old memory sits at $t \approx (10/11)^{30} \approx 0.047$, deep in the leftward portion of the protocol where rebinning-induced blurring is most severe.

\subsection{Parameter dependence}

\begin{figure}[h!]
\centering
\includegraphics[width=0.95\linewidth]{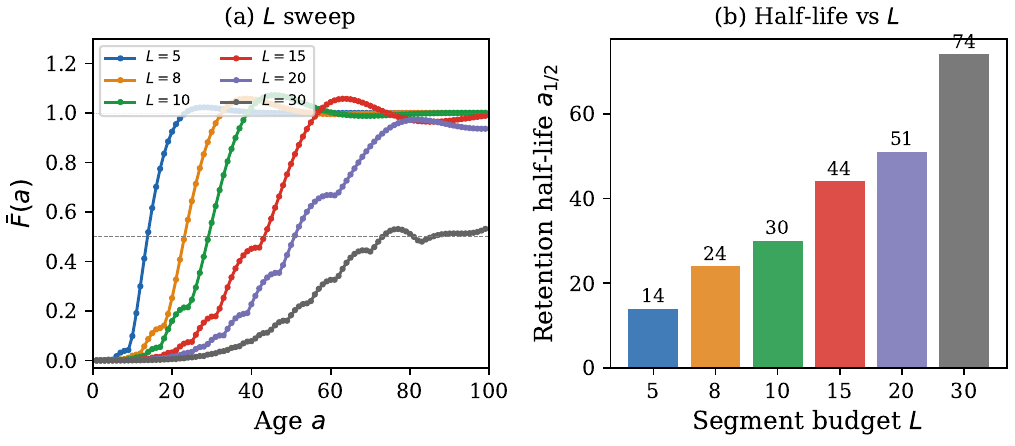}
\caption{Segment-budget sweep for the single-Gaussian circular-drift experiment.  (a)~Age--forgetting curves for $L \in \{5, 8, 10, 15, 20, 30\}$.  Increasing~$L$ shifts the sigmoid transition to higher ages without changing the curve shape qualitatively.  (b)~Retention half-life $a_{1/2}$ versus~$L$, confirming approximate linear scaling.}
\label{fig:k1_L_sweep}
\end{figure}

The segment budget~$L$ is the primary determinant of retention.  Sweeping $L \in \{5, 8, 10, 15, 20, 30\}$ yields half-lives $a_{1/2} \in \{14, 24, 30, 44, 51, 74\}$, scaling roughly as $a_{1/2} \approx 2.4\,L$ (Fig.~\ref{fig:k1_L_sweep}).  This near-linear scaling is consistent with the observation that each CAS cycle degrades the readout time by a factor $L/(L+1)$, so the number of cycles before a memory reaches a fixed resolution threshold is proportional to~$L$.

Drift speed modulates the half-life: faster drift (shorter period~$P$) leads to shorter retention, because larger daily displacements accumulate more error through rebinning.  Sweeping $P \in \{25, 50, 100, 200\}$ yields $a_{1/2} \in \{20, 30, 34, 36\}$.  The dependence saturates for slow drift ($P \ge 100$), suggesting a floor set by the diffusive contribution of the rebinning itself.

Drift geometry (circular vs.\ linear) affects the curve shape more than the half-life: linear drift yields a clean monotone sigmoid with $a_{1/2} = 42$ (vs.\ $30$ for circular at the same $L$), while circular drift introduces non-monotone modulations due to periodic spatial recurrence.  The higher linear-drift half-life reflects the absence of geometric aliasing: each recalled location is unique, so the rebinning error is always genuine.

\begin{figure}[h!]
\centering
\includegraphics[width=0.95\linewidth]{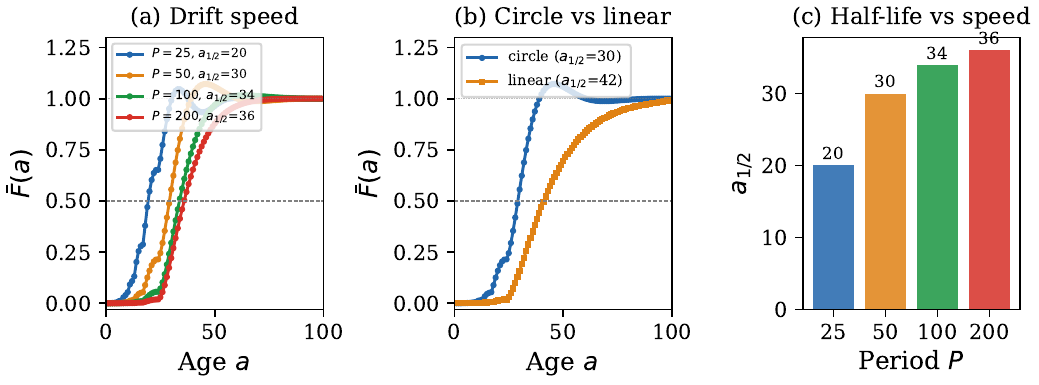}
\caption{(a)~Drift-speed sweep (circle period $P$).  Faster drift shortens the half-life, but the effect saturates for slow drift.  (b)~Circle vs.\ linear drift.  The circular curve is non-monotone due to periodic spatial recurrence; the linear curve is a clean sigmoid with longer half-life.  (c)~Half-life versus period~$P$.}
\label{fig:k1_speed_shape}
\end{figure}

\subsection{Takeaway}

The $K=1$ experiments establish three main results.  First, the forgetting curve has a universal two-regime shape (plateau $+$ sigmoid) whose transition is controlled by the segment budget~$L$, with $a_{1/2} \approx 2.4\,L$.  This is the first observation of the linear retention-capacity law, which we will confirm across progressively more complex settings: multi-component mixtures (Section~\ref{sec:phase2}), crowding and dimension scaling (Section~\ref{sec:scaling}), and image-derived latent spaces (Section~\ref{sec:mnist_latent}).  Second, drift speed modulates the half-life but drift geometry affects only the curve shape.  Third, forgetting manifests as confusion (displacement toward recent eras), not destruction (reversion to the prior).  These findings motivate the $K>1$ experiments below, where we test whether state-space complexity affects the retention timescale.

\section{Experiments: Gaussian mixtures ($K=3$, $d=2$)}
\label{sec:phase2}

We now extend to $K$-component Gaussian-mixture daily targets.  Each day's distribution has $K=3$ equal-weight components arranged in a rotating equilateral triangle of radius $r=0.8$ around the drifting circle centre (same circular drift as Section~\ref{sec:phase1}, with per-component covariance $0.3\,I$).

\subsection{Default run and decomposed forgetting}

\begin{figure}[h!]
\centering
\includegraphics[width=0.95\linewidth]{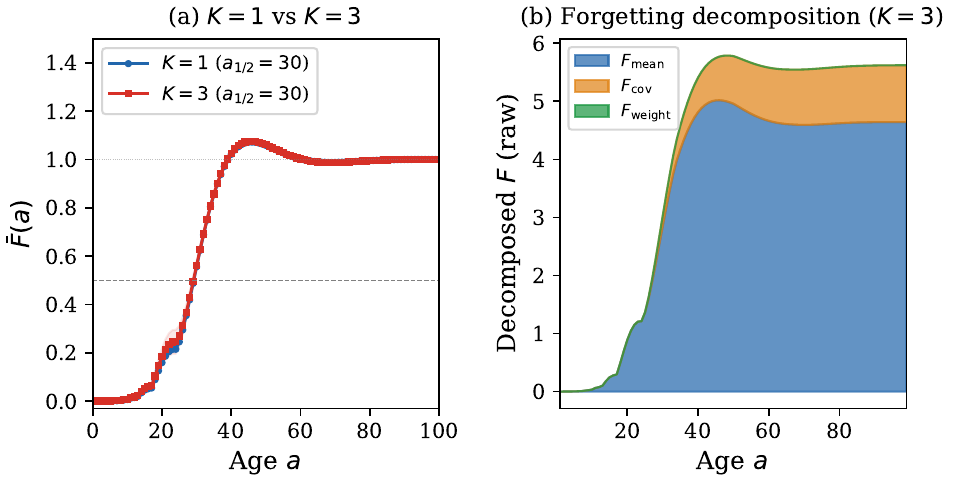}
\caption{(a)~Age--forgetting curves for $K=1$ (blue) and $K=3$ (red), both with $L=10$ and circular drift.  The curves nearly coincide, with both yielding $a_{1/2} = 30$.  (b)~Decomposed raw forgetting for $K=3$: mean misalignment (blue) dominates, covariance error (orange) is secondary, and weight error (green) is negligible.}
\label{fig:k3_comparison}
\end{figure}

With $L=10$, the $K=3$ experiment yields $a_{1/2} = 30$ --- \emph{identical} to the $K=1$ case (Fig.~\ref{fig:k3_comparison}a).  This is the first indication that retention is governed by the temporal budget~$L$ rather than the state-space complexity~$K$.

The decomposed forgetting (Fig.~\ref{fig:k3_comparison}b) reveals that $F_{\rm mean}$ dominates (${\sim}85\%$ of total raw forgetting), $F_{\rm cov}$ contributes ${\sim}15\%$, and $F_{\rm weight}$ is negligible (of order $10^{-17}$, i.e.\ machine precision).  The vanishing weight error is a structural consequence of equal-weight mixtures: convex combinations of equal weights remain equal, so the rebinning preserves weights exactly.

\subsection{Component-level trajectories}

\begin{figure}[h!]
\centering
\includegraphics[width=0.95\linewidth]{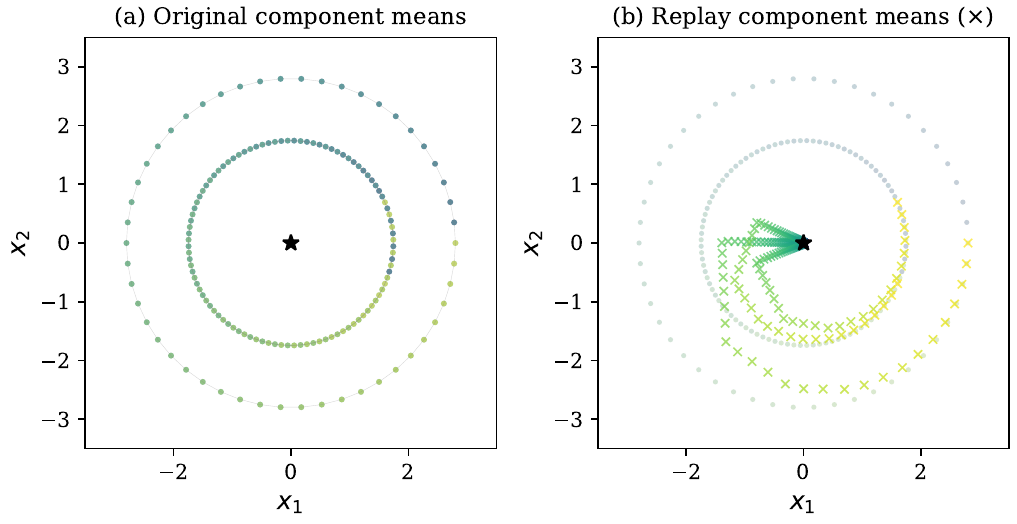}
\caption{Component-level trajectories for $K=3$.  (a)~Original daily component means, coloured by day.  The three interleaved helical paths trace the rotating-triangle geometry.  (b)~Replayed component means ($\times$) at the final day, after Hungarian matching to the originals.  Recent components are well-recalled; older ones collapse toward the origin.}
\label{fig:k3_trajectories}
\end{figure}

Fig.~\ref{fig:k3_trajectories} shows the per-component replayed means (after Hungarian matching) at the final day.  Recent days' component means are replayed accurately; older days collapse toward the protocol interior, with all three components converging toward a common cluster near the origin.  This mirrors the $K=1$ confusion pattern, amplified by the need to simultaneously track three interacting trajectories.

\subsection{$L$ sweep and $K$ sweep}

\begin{figure}[h!]
\centering
\includegraphics[width=0.95\linewidth]{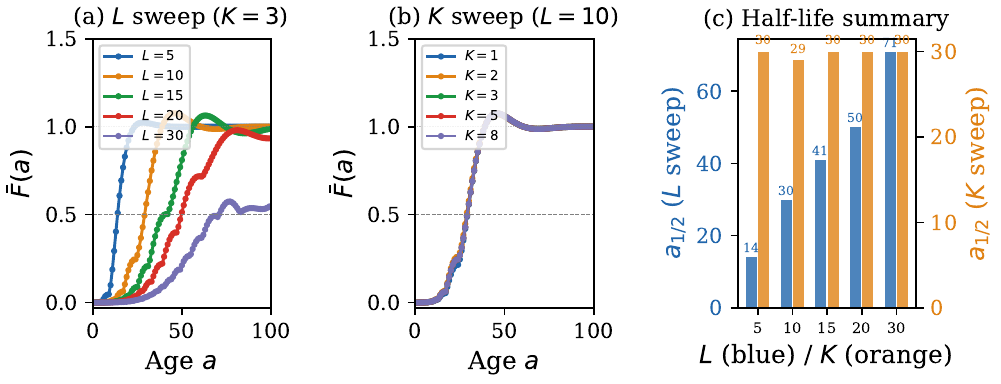}
\caption{(a)~$L$ sweep at $K=3$: age--forgetting curves for $L \in \{5, 10, 15, 20, 30\}$.  The pattern mirrors the $K=1$ case.  (b)~$K$ sweep at $L=10$: age curves for $K \in \{1, 2, 3, 5, 8\}$ nearly coincide.  (c)~Half-life summary for both sweeps.}
\label{fig:k3_sweeps}
\end{figure}

Sweeping the segment budget at $K=3$ gives half-lives $a_{1/2} \in \{14, 30, 41, 50, 71\}$ for $L \in \{5, 10, 15, 20, 30\}$ (Fig.~\ref{fig:k3_sweeps}a), closely matching the $K=1$ results.

The $K$ sweep at $L=10$ (Fig.~\ref{fig:k3_sweeps}b) yields $a_{1/2} \in \{30, 29, 30, 30, 30\}$ for $K \in \{1, 2, 3, 5, 8\}$ --- the half-life is essentially \emph{flat} across mixture complexity.  This is the paper's central experimental finding: \emph{retention is controlled by the temporal budget~$L$, not by the state-space complexity~$K$}.  The $K$-independence is not approximate: the half-life varies by at most one day across a factor-of-8 range in~$K$.

\subsection{Takeaway}

The $K>1$ experiments establish two main results.  First, the half-life is independent of~$K$: adding mixture components does not shorten (or lengthen) retention.  This is because the rebinning step treats all GM parameters (weights, means, covariances) uniformly --- the interpolation does not ``see'' how many components there are.  Second, forgetting is overwhelmingly driven by mean misalignment; covariance error is secondary and weight error is negligible for equal-weight mixtures.  These findings justify using the half-life~$a_{1/2}$ as a single scalar summary of retention quality, controlled by~$L$ alone.

\section{Scaling experiments}
\label{sec:scaling}

We now test how the continual memory mechanism scales when the daily targets become more crowded, when the relevant signal is embedded into a higher-dimensional ambient space, and when the target family undergoes a simple topological curriculum involving split-and-merge events. The goal of this section is not to optimize performance, but to identify which aspects of increasing problem complexity actually shorten retention and which do not.

Based on the $K$-independence result of Section~\ref{sec:phase2} --- specifically, the flat half-life across $K \in \{1,2,3,5,8\}$ --- we conjectured that the same qualitative picture would persist: forgetting is governed primarily by temporal compression under a fixed protocol budget, while many forms of static state-space complexity affect the geometry of replay far more than the retention timescale itself.  The experiments below confirm this conjecture under three increasingly challenging scenarios.

\subsection{Crowding as a control parameter}

\begin{figure}[h!]
\centering
\includegraphics[width=0.98\linewidth]{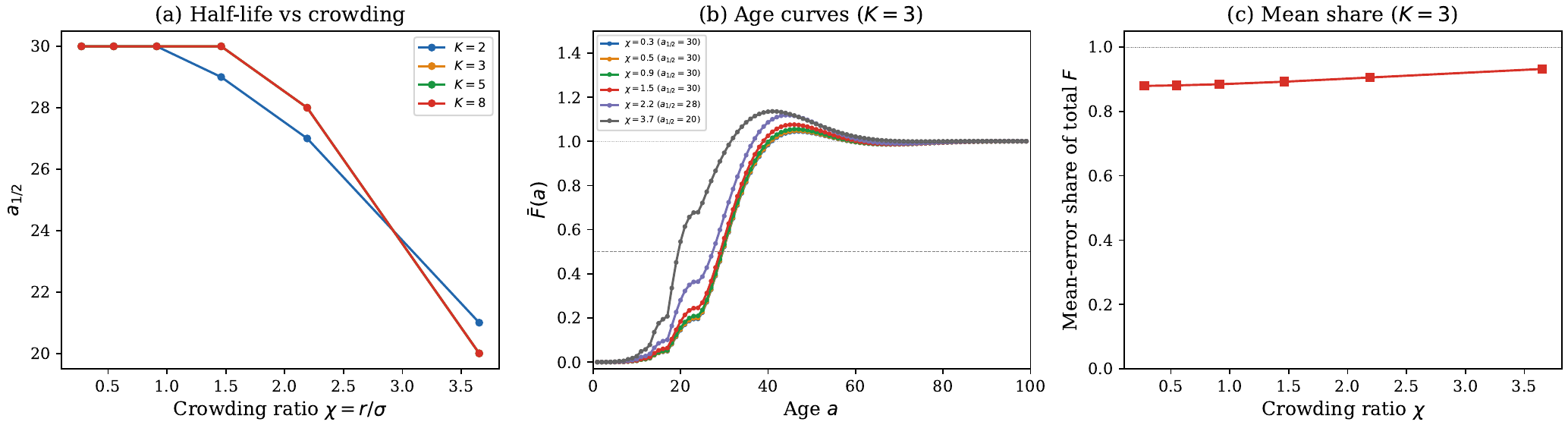}
\caption{Crowding sweep for Gaussian-mixture memories ($L=10$). (a)~Retention half-life $a_{1/2}$ versus crowding ratio $\chi$ for $K=2,3,5,8$.  The half-life is flat at~30 for $\chi \lesssim 1.5$ and declines only for well-separated components.  (b)~Age--forgetting curves at $K=3$ for six crowding values.  (c)~Mean-error share of total forgetting, stable at ${\sim}90\%$ across all~$\chi$.
}
\label{fig:scaling_crowding}
\end{figure}

We begin with mixtures in $d=2$, varying the \emph{crowding ratio} $\chi = r/\sigma$, where $r$ is the inter-component offset radius and $\sigma = \sqrt{\mathrm{cov\_scale}}$ is the component standard deviation.  Small~$\chi$ corresponds to heavily overlapping components (strong crowding); large~$\chi$ to well-separated components (weak crowding).  We sweep $r \in \{0.15, 0.3, 0.5, 0.8, 1.2, 2.0\}$ at $K=3$, corresponding to $\chi \in \{0.27, 0.55, 0.91, 1.46, 2.19, 3.65\}$.  Fig.~\ref{fig:scaling_crowding} summarizes the results.

Panel~(a) shows retention half-life $a_{1/2}$ versus crowding ratio for $K=2,3,5,8$.  The first observation is that all curves are \emph{flat at $a_{1/2}=30$} for $\chi \lesssim 1.5$: moderate-to-strong crowding has no effect on retention whatsoever.  Only at high separation ($\chi > 2$) does the half-life begin to decrease, dropping to $a_{1/2} \approx 20$ at $\chi = 3.65$.  The effect is most pronounced for $K=2$, whose half-life begins declining earlier (at $\chi \approx 1$) than for $K \ge 3$.  This decline at large~$\chi$ is a geometric effect: when components are widely separated, each component's mean displacement under rebinning is larger in absolute terms, accelerating forgetting.

Panel~(b) shows the age--forgetting curves at $K=3$ for six crowding values.  The curves for $\chi \le 1.5$ are nearly indistinguishable, all showing the standard sigmoid with $a_{1/2}=30$.  At $\chi = 2.2$ the half-life shortens slightly to~28, and at $\chi = 3.7$ it drops to~20, with the sigmoid onset shifting leftward and the confusion overshoot ($\bar F > 1$) increasing.

Panel~(c) reports the average share of raw forgetting attributable to mean misalignment.  The mean-error share is ${\sim}90\%$ across all crowding ratios, confirming that even as crowding changes the spatial geometry of forgetting, the dominant error channel remains mean displacement rather than covariance distortion or weight drift.

\begin{figure}[h!]
\centering
\includegraphics[width=0.88\linewidth]{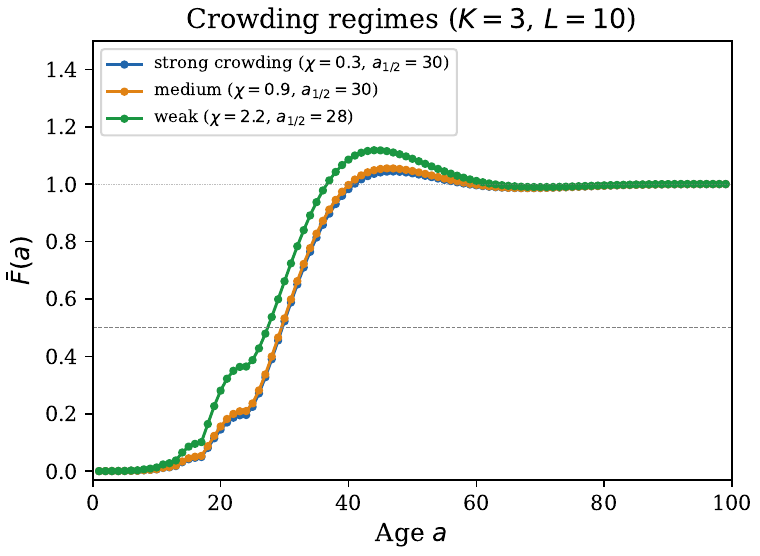}
\caption{Representative age-averaged forgetting curves across three crowding regimes ($K=3$, $L=10$).  Strong ($\chi=0.3$) and medium ($\chi=0.9$) crowding yield indistinguishable curves ($a_{1/2}=30$).  Weak crowding ($\chi=2.2$) produces a slightly earlier transition ($a_{1/2}=28$) and a more pronounced confusion overshoot.
}
\label{fig:scaling_agecurves}
\end{figure}

To illustrate the shape of forgetting more directly, Fig.~\ref{fig:scaling_agecurves} shows three representative age-forgetting curves corresponding to strong ($\chi = 0.3$), medium ($\chi = 0.9$), and weak ($\chi = 2.2$) crowding.  The most notable feature is that the strong and medium crowding curves are virtually identical ($a_{1/2} = 30$ for both), while the weakly crowded case shows a slightly earlier sigmoid onset ($a_{1/2} = 28$) and a more pronounced confusion overshoot ($\bar F \approx 1.12$ vs.\ $\approx 1.05$).  The overshoot amplification at weak crowding is consistent with larger per-component mean displacements when components are far apart.

\subsection{Fixed low-dimensional signal in a higher-dimensional ambient space}

\begin{figure}[h!]
\centering
\includegraphics[width=0.98\linewidth]{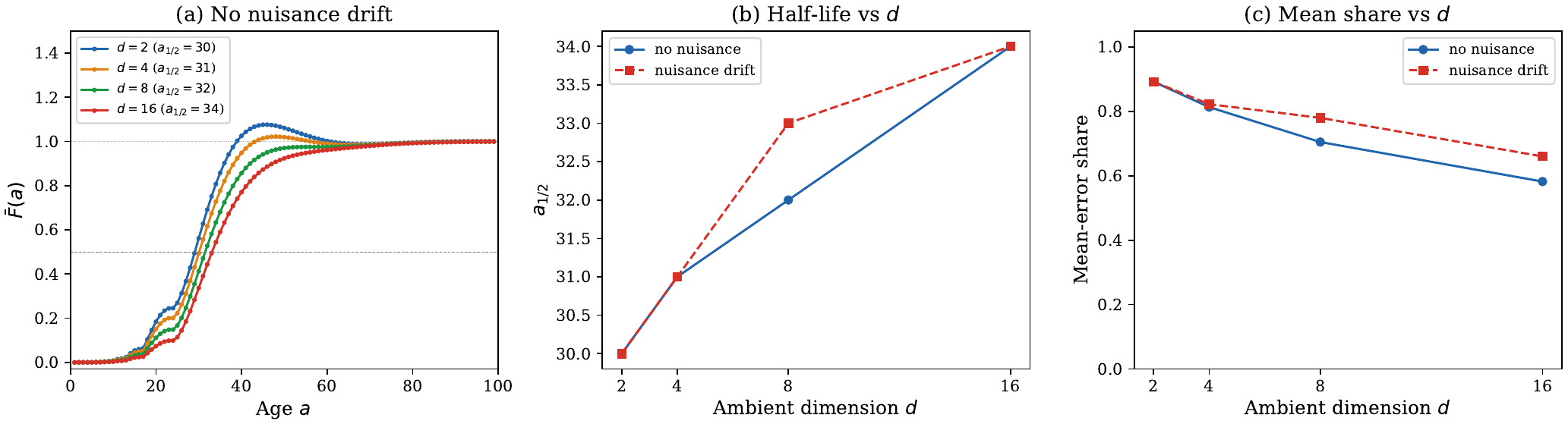}
\caption{
Scaling with ambient dimension for a fixed 2D signal ($K=3$, $L=10$).  (a)~Age-forgetting curves for $d=2,4,8,16$ without nuisance drift.  Higher~$d$ slightly \emph{improves} normalised retention due to the larger amnesia baseline.  (b)~Retention half-life versus ambient dimension under two nuisance settings.  Both show gentle increase with~$d$.  (c)~Mean-error share decreases with~$d$ as covariance mismatch grows in the extra dimensions.
}
\label{fig:scaling_dimension}
\end{figure}

We next test whether retention degrades when the informative signal remains two-dimensional but is embedded into a higher-dimensional ambient space.  Fig.~\ref{fig:scaling_dimension} reports the results for ambient dimensions $d=2,4,8,16$, with $K=3$ and $L=10$.

Panel~(a) shows the age-forgetting curves when the extra dimensions carry no drift (nuisance coordinates remain at zero).  The curves shift \emph{rightward} with increasing~$d$: the half-life increases slightly from $a_{1/2}=30$ at $d=2$ to $a_{1/2}=34$ at $d=16$.  This counter-intuitive improvement occurs because the amnesia baseline $F_{\rm amnesia}$ grows with~$d$ (the prior covariance is $I_d$, contributing more Frobenius-norm distance from each daily target), while the rebinning error in the signal subspace is unchanged.  The normalised forgetting is therefore diluted by the larger baseline.

Panel~(b) summarizes the half-life as a function of~$d$ for two settings.  When nuisance dimensions are static, $a_{1/2}$ increases gently from~30 to~34.  When nuisance dimensions carry a slow random walk (speed~0.1/day), the half-life follows a similar trend ($a_{1/2} \approx 30$--$33$), indicating that moderate nuisance drift does not substantially impair retention of the signal.

Panel~(c) shows that the mean-error share declines with~$d$, from ${\sim}90\%$ at $d=2$ to ${\sim}60\%$ at $d=16$ in the no-nuisance setting.  This shift reflects the growing contribution of covariance mismatch in the extra dimensions: as~$d$ increases, the $d \times d$ covariance matrices carry more entries that can accumulate rebinning error.  With nuisance drift, the mean-error share remains higher (${\sim}70\%$ at $d=16$) because the drifting nuisance means contribute additional mean-channel error.

\subsection{Split-and-merge curriculum}

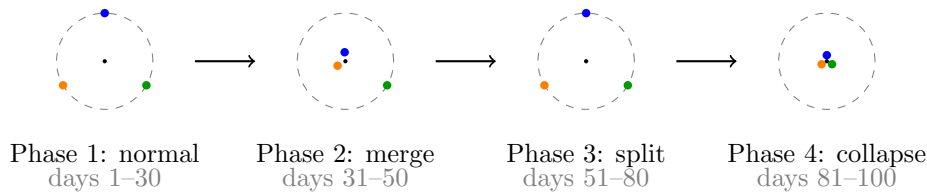
\begin{figure}[t]
\centering
\begin{tikzpicture}[scale=0.8, every node/.style={font=\small}]
  \begin{scope}[shift={(0,0)}]
    \draw[gray,dashed] (0,0) circle (0.8);
    \fill[blue] (90:0.8) circle (2pt);
    \fill[orange] (210:0.8) circle (2pt);
    \fill[green!60!black] (330:0.8) circle (2pt);
    \fill (0,0) circle (1pt);
    \node[below] at (0,-1.2) {Phase 1: normal};
    \node[below,gray] at (0,-1.6) {days 1--30};
  \end{scope}
  \draw[->,thick] (1.5,0) -- (2.5,0);
  \begin{scope}[shift={(4,0)}]
    \draw[gray,dashed] (0,0) circle (0.8);
    \fill[blue] (95:0.15) circle (2pt);
    \fill[orange] (210:0.15) circle (2pt);
    \fill[green!60!black] (330:0.8) circle (2pt);
    \fill (0,0) circle (1pt);
    \node[below] at (0,-1.2) {Phase 2: merge};
    \node[below,gray] at (0,-1.6) {days 31--50};
  \end{scope}
  \draw[->,thick] (5.5,0) -- (6.5,0);
  \begin{scope}[shift={(8,0)}]
    \draw[gray,dashed] (0,0) circle (0.8);
    \fill[blue] (90:0.8) circle (2pt);
    \fill[orange] (210:0.8) circle (2pt);
    \fill[green!60!black] (330:0.8) circle (2pt);
    \fill (0,0) circle (1pt);
    \node[below] at (0,-1.2) {Phase 3: split};
    \node[below,gray] at (0,-1.6) {days 51--80};
  \end{scope}
  \draw[->,thick] (9.5,0) -- (10.5,0);
  \begin{scope}[shift={(12,0)}]
    \draw[gray,dashed] (0,0) circle (0.8);
    \fill[blue] (90:0.1) circle (2pt);
    \fill[orange] (210:0.1) circle (2pt);
    \fill[green!60!black] (330:0.1) circle (2pt);
    \fill (0,0) circle (1pt);
    \node[below] at (0,-1.2) {Phase 4: collapse};
    \node[below,gray] at (0,-1.6) {days 81--100};
  \end{scope}
\end{tikzpicture}
\caption{Schematic of the four-phase split-and-merge curriculum for $K=3$.  Coloured dots represent the three mixture component means; the dashed circle indicates the inter-component radius~$r$.  Phase~1: normal rotating triangle ($r=0.8$).  Phase~2: two components merge ($r_{01}\to 0.05$).  Phase~3: split back to triangle ($r\to 0.8$).  Phase~4: all three collapse toward the centre ($r\to 0.1$).  Transitions are smoothed over 5-day ramps.}
\label{fig:split_merge_scheme}
\end{figure}

\begin{figure}[h!]
\centering
\includegraphics[width=0.98\linewidth]{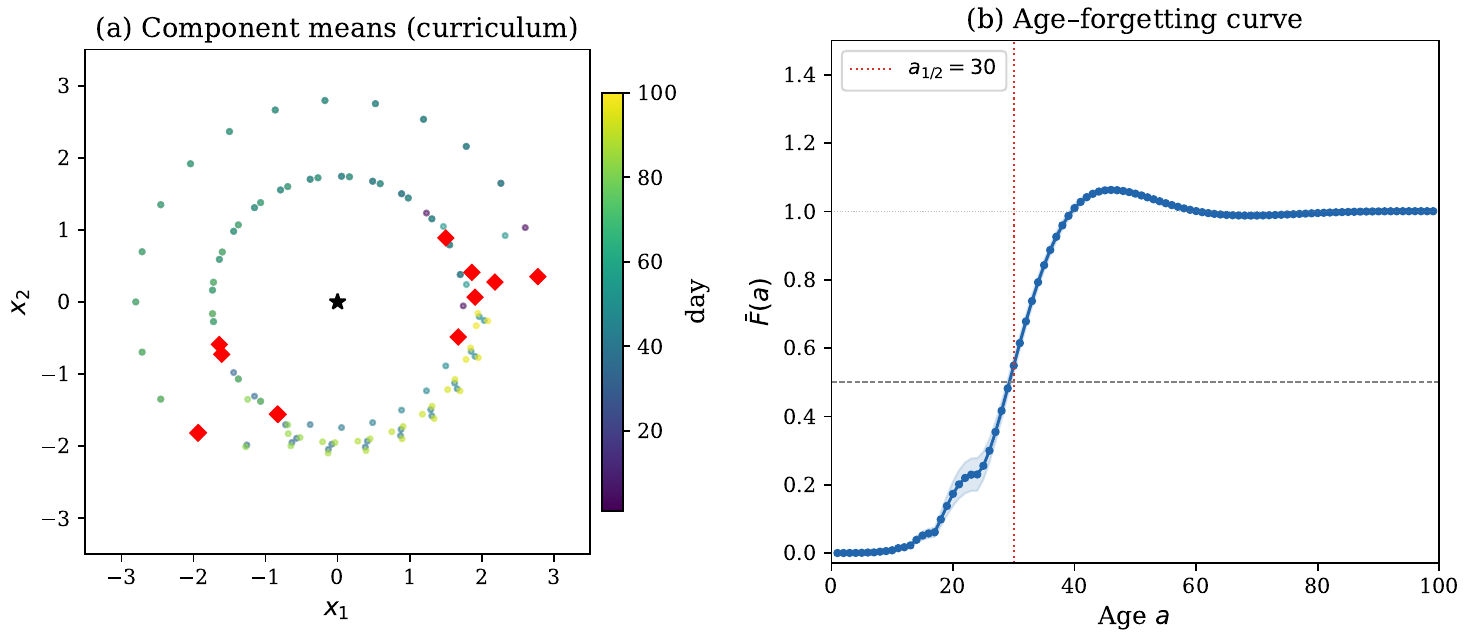}
\caption{
Split-and-merge curriculum experiment ($K=3$, $L=10$).  (a)~Daily component means, coloured by day, with phase-boundary markers (red diamonds) at days 1, 31, 51, and~81.  The four phases --- normal triangle, merge, split, collapse --- are clearly visible.  (b)~Age-averaged forgetting curve.  Despite the nontrivial curriculum, the retention half-life remains $a_{1/2}=30$, identical to the stationary baseline.
}
\label{fig:scaling_curriculum}
\end{figure}

As a final scaling test, we consider a simple curriculum in which the daily mixture geometry changes topologically over time via split-and-merge events.  The $K=3$ mixture undergoes four phases, illustrated schematically in Fig.~\ref{fig:split_merge_scheme}: a normal rotating triangle ($r=0.8$, days 1--30), a merge phase where two components collapse toward each other ($r_{01} \to 0.05$, days 31--50), a split phase where they separate again ($r \to 0.8$, days 51--80), and a final collapse where all three components converge toward the centre ($r \to 0.1$, days 81--100).  Transitions are smoothed over 5-day ramps.  Fig.~\ref{fig:scaling_curriculum} shows both the daily component means and the resulting age-forgetting curve.

Panel~(a) displays the component centres across the 100 days, with phase-boundary markers (red diamonds) at days 1, 31, 51, and~81.  The four phases are clearly visible: the initial rotating triangle, the merged pair, the re-separation, and the final collapse.

Panel~(b) shows the corresponding age-forgetting curve.  Despite the nontrivial topological evolution, the half-life is $a_{1/2} = 30$ --- identical to the stationary-geometry baseline.  The curve shape is the standard sigmoid with a mild non-monotone feature around age $10$--$15$, attributable to the interaction between curriculum transitions and the periodic drift geometry.

This result is the strongest evidence that \emph{the retention timescale is set by the temporal budget~$L$ alone}: even when the daily target distribution undergoes qualitative structural changes --- merging, splitting, and collapsing of mixture components --- the half-life is unaffected.

\subsection{Overview and interpretation}

The three scaling experiments paint a consistent picture.  Crowding affects the half-life only at extreme separation ($\chi > 2$, where per-component mean displacements become large), and even then the reduction is modest (from~30 to~20).  Ambient dimension either has no effect or slightly \emph{improves} normalised retention (due to the growing amnesia baseline), while shifting the forgetting channel from mean-dominated toward a more even mean/covariance split.  A time-varying curriculum with topological changes leaves the half-life entirely unchanged.

These results confirm the conjecture from Section~\ref{sec:phase2}: the retention half-life $a_{1/2} \approx 2.4\,L$ is a robust, universal characteristic of the CAS recursion under uniform-grid rebinning.  It depends on the temporal budget~$L$ and, to a lesser extent, on the drift speed, but is insensitive to the state-space complexity~$K$, the ambient dimension~$d$, the crowding geometry, and even topological changes in the daily target family.  The only avenue for substantially improving retention, within the current framework, is to increase~$L$ or to replace the uniform grid with an adaptive one that allocates finer temporal resolution to recent memories.

\section{MNIST latent-space illustration}
\label{sec:mnist_latent}

To complement the analytically controlled Gaussian-mixture experiments, we construct an image-based latent-space illustration using MNIST.  The purpose is twofold: (i)~to test whether the same notions of age-dependent forgetting, confusion, and retention-time control carry over when the GM components represent real image classes; and (ii)~to demonstrate the ``movie'' capability described in Section~\ref{sec:stochastic_replay} --- the protocol grid, decoded frame-by-frame to pixel space, produces a visual temporal narrative of the agent's compressed history.

\subsection{Setup: latent embeddings and rotating-dominance curriculum}

\begin{figure}[h!]
\centering
\includegraphics[width=0.8\linewidth]{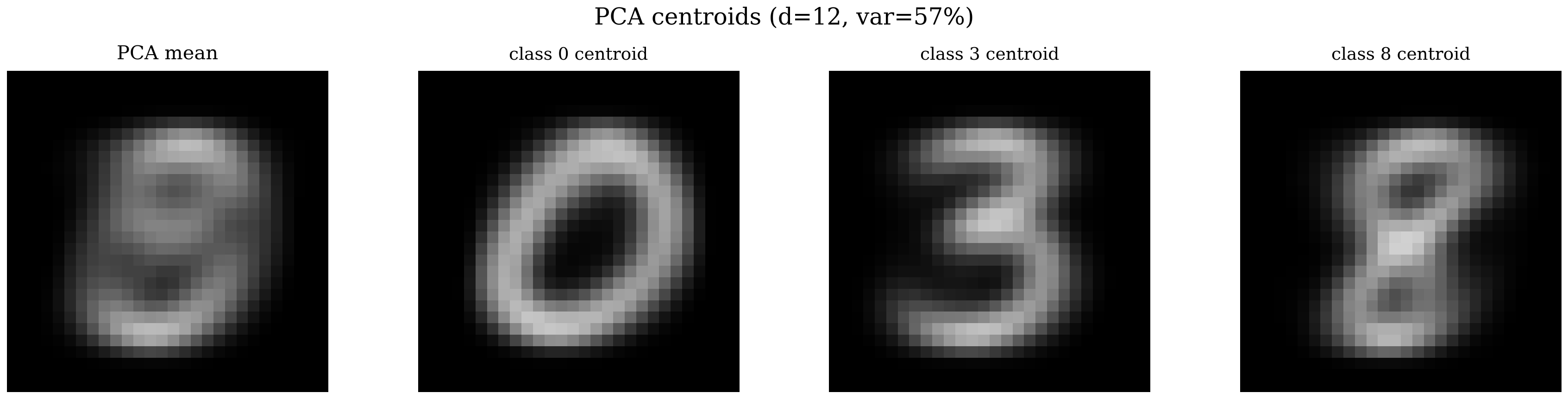}
\caption{PCA centroids at $d=12$: the global mean (left) and per-class centroids for digits 0, 3, and~8.  Despite capturing only $57\%$ of total variance, the centroids are visually recognisable.}
\label{fig:mnist_centroids}
\end{figure}

We select three visually distinct MNIST digit classes --- $0$, $3$, and $8$ --- and embed the corresponding ${\sim}18{,}000$ training images into a $d=12$ PCA latent space ($57\%$ explained variance).  At this dimension, PCA-decoded class centroids are clearly recognisable as their respective digits (Fig.~\ref{fig:mnist_centroids}).

We fit a single Gaussian per class ($K=3$ total) and construct a \emph{rotating-dominance curriculum} over $n=100$ days: the component means and covariances are fixed to their class-conditional fits, while the mixing weights rotate with period $P=30$:
\begin{equation}
  \pi_k^{(m)} = \mathrm{softmax}\bigl(A\,\cos(2\pi m/P + 2\pi k/3)\bigr),
  \qquad A=2,
\end{equation}
so that each digit class cycles between dominance ($\pi_k\approx 0.9$) and near-absence ($\pi_k\approx 0.04$).  This is the semantic analogue of the synthetic circular drift: the ``location'' in distribution space rotates through digit classes rather than through spatial coordinates.

\subsection{Forgetting curve and comparison with synthetic experiments}

\begin{figure}[h!]
\centering
\includegraphics[width=0.55\linewidth]{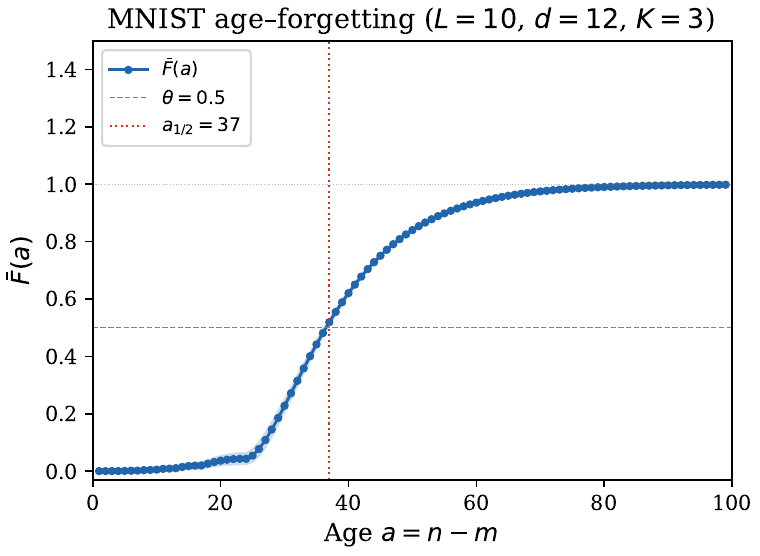}
\caption{MNIST age--forgetting curve ($L=10$, $d=12$, $K=3$).  The half-life $a_{1/2}=37$ and the curve shape is a clean sigmoid without the confusion overshoot seen in the synthetic mean-drift experiments.}
\label{fig:mnist_age_curve}
\end{figure}

Running CAS with $L=10$ yields a retention half-life of $a_{1/2}=37$ (Fig.~\ref{fig:mnist_age_curve}).  The age--forgetting curve exhibits the familiar two-regime structure --- a low-error plateau followed by a sigmoid transition --- with no confusion overshoot ($\bar F$ saturates at~1 rather than exceeding it).  The absence of overshoot is explained by the nature of the daily variation: since only the weights change (not the component means), the replayed means for old days converge toward a time-averaged centroid rather than being actively displaced past it.

\begin{figure}[h!]
\centering
\includegraphics[width=0.55\linewidth]{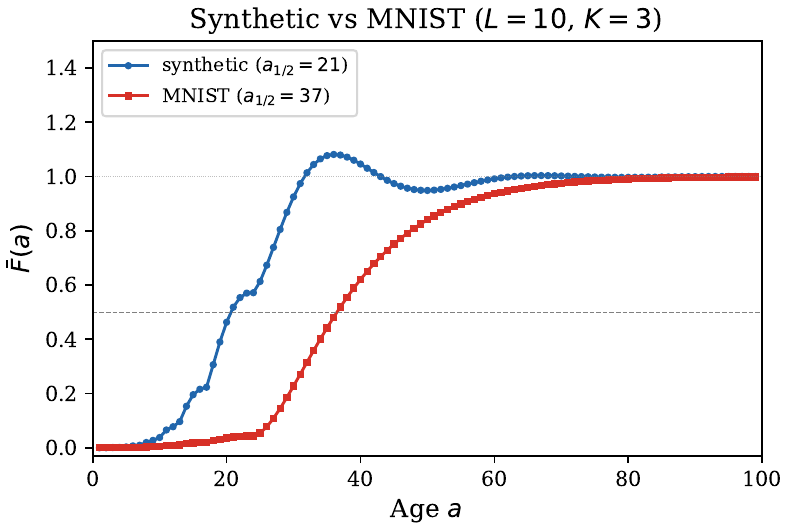}
\caption{Comparison of synthetic $K=3$ (blue, $d=2$, circle drift, $P=30$) and MNIST (red, $d=12$, rotating-dominance weights, $P=30$) age--forgetting curves at $L=10$.  The MNIST curve is shifted rightward due to the higher dimension and milder daily perturbation.}
\label{fig:mnist_comparison}
\end{figure}

Fig.~\ref{fig:mnist_comparison} compares the MNIST and synthetic $K=3$ forgetting curves at the same $L=10$ and $P=30$.  The MNIST half-life ($a_{1/2}=37$) exceeds the synthetic one ($a_{1/2}=21$).  Two effects contribute: (i)~the higher latent dimension $d=12$ inflates the amnesia baseline, diluting the normalised metric; and (ii)~the MNIST curriculum perturbs only the weights (a $K$-dimensional vector), while the synthetic curriculum moves all $K$ component means through $\R^2$, producing larger per-step rebinning error.

\subsection{Decomposed forgetting: covariance-dominated regime}

\begin{figure}[h!]
\centering
\includegraphics[width=0.65\linewidth]{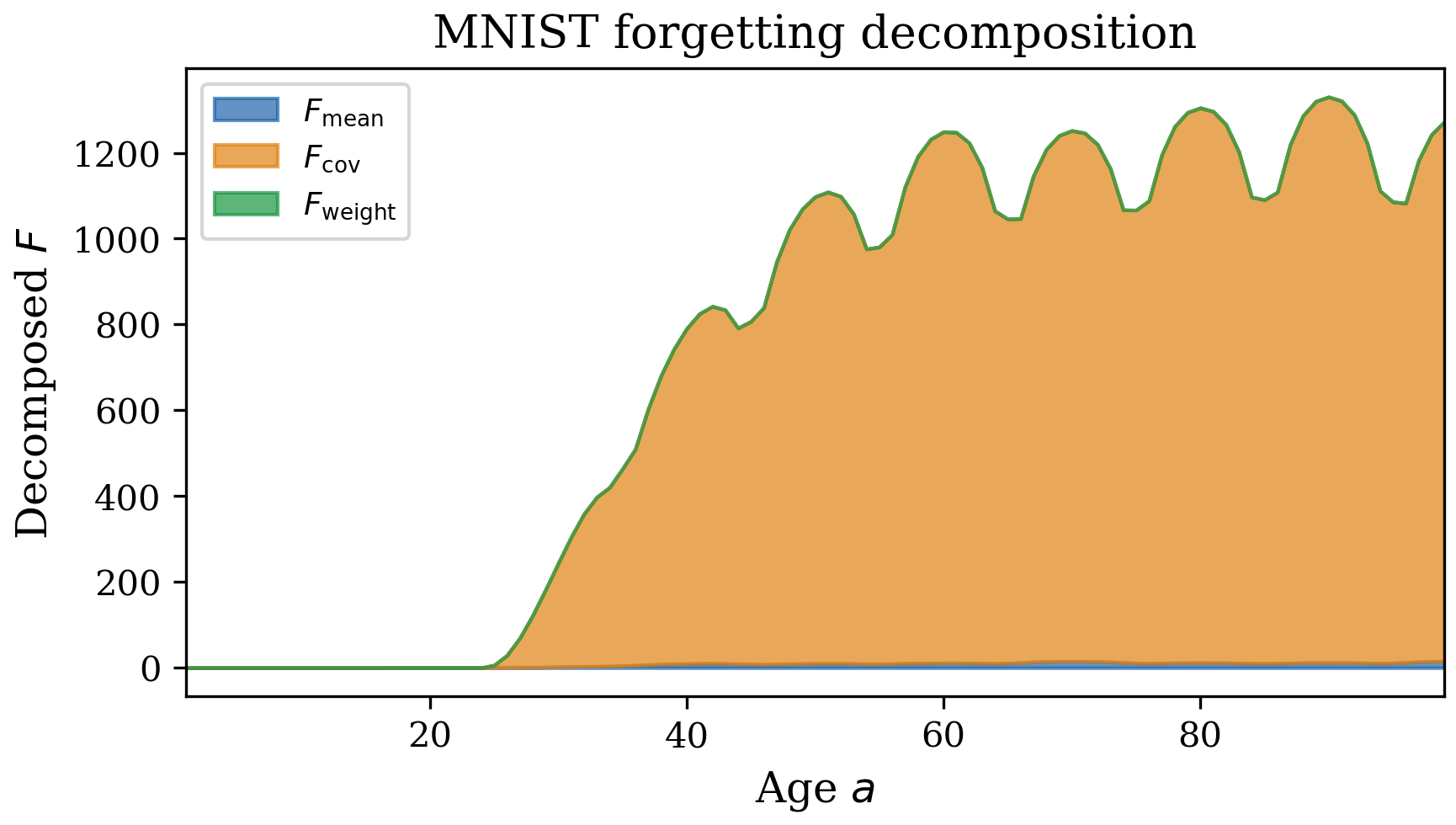}
\caption{Decomposed raw forgetting for MNIST.  Unlike the synthetic experiments where $F_{\rm mean}$ dominates, the MNIST construction is covariance-dominated because component means are fixed and only the weights rotate.  The periodic oscillations at large age have period~$P=30$, matching the curriculum: dips occur at ages that are multiples of~$P$ (where the recalled and current days share the same dominant digit class), while peaks occur at half-period offsets.}
\label{fig:mnist_decomposed}
\end{figure}

The forgetting decomposition (Fig.~\ref{fig:mnist_decomposed}) reveals a qualitative difference from the synthetic experiments.  In the MNIST construction, $F_{\rm cov}$ dominates the raw forgetting, accounting for the overwhelming majority of the total, while $F_{\rm mean}$ is comparatively small and $F_{\rm weight}$ contributes visibly but remains secondary.  This is the opposite of the synthetic case (where $F_{\rm mean}\approx 85\%$) and is explained by the design of the curriculum: since component means are fixed, the mean channel accumulates minimal rebinning drift; instead, the $d\times d$ covariance matrices ($12\times 12 = 144$ entries per component) accumulate Frobenius-norm error as the piecewise-linear interpolation progressively distorts the class-specific covariance structure.  The weight error is non-negligible here because the rotating weights are the primary information channel, unlike the synthetic equal-weight setting.

The raw forgetting also exhibits periodic oscillations at large age (visible in Fig.~\ref{fig:mnist_decomposed}), with period~$P=30$ matching the curriculum.  The mechanism is straightforward: with $K=3$ classes cycling with period~$P=30$, days separated by exactly $30$ (or $60$, $90$, \ldots) share the same dominant digit class.  When such a day is recalled, the replayed weight vector happens to be closer to the original (since both have the same class dominant), producing a dip in raw forgetting.  Days at half-period offsets ($a=15, 45, \ldots$) have maximally mismatched weight vectors and produce forgetting peaks.  This resonance effect is purely a consequence of the periodic curriculum and has no analogue in the synthetic mean-drift experiments, where the daily dynamics are continuous rather than weight-modulated.

\subsection{Visual forgetting and the temporal movie}

The key diagnostic of the MNIST experiment is visual: decoded images reveal how forgetting manifests in pixel space.

\begin{figure}[h!]
\centering
\includegraphics[width=0.95\linewidth]{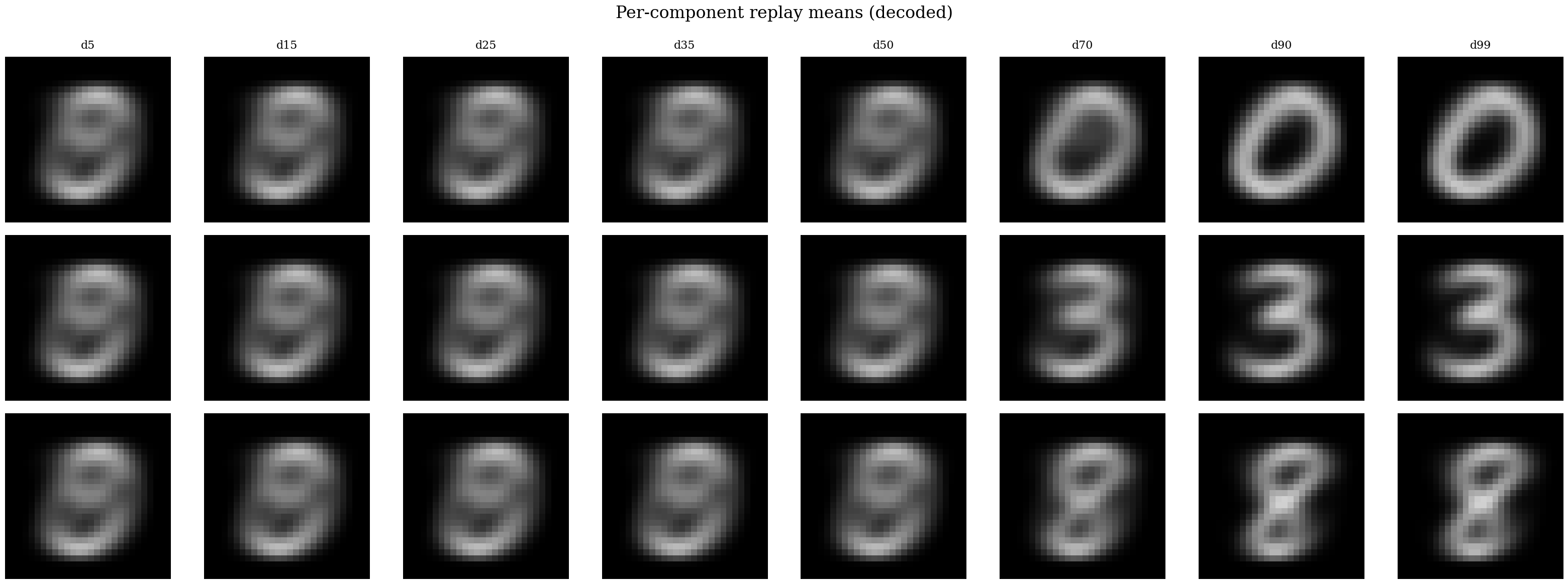}
\caption{Per-component replayed means decoded to pixel space, for selected past days.  Each row corresponds to one digit class (0, 3, 8 from top to bottom).  Recent days show distinct digit identities; older days progressively converge toward a common blurred average.}
\label{fig:mnist_visual}
\end{figure}

Fig.~\ref{fig:mnist_visual} shows the per-component replayed means (decoded to $28\!\times\!28$ pixels) for eight selected past days.  For recent days (d90, d99), the three components decode to clearly distinct, recognisable digits (0, 3, 8).  As age increases, the components blur and converge: by day~25 and earlier, all three components decode to a similar ambiguous shape resembling the PCA grand mean.  This is the visual manifestation of confusion --- not semantic collapse (which would mean instant failure), but progressive loss of class identity through cumulative rebinning.

\begin{figure}[h!]
\centering
\includegraphics[width=0.98\linewidth]{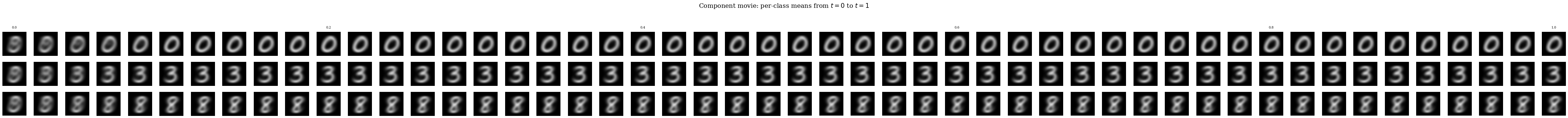}
\caption{Component movie: per-class decoded means from $t=0$ (oldest memories, left) to $t=1$ (current day, right).  Each row is one digit class.  Digit identities are preserved across the full protocol, with only gradual loss of sharpness at the oldest end.}
\label{fig:mnist_movie}
\end{figure}

The protocol grid, evaluated at uniformly spaced times $t\in[0,1]$ and decoded frame-by-frame, produces a \emph{temporal movie} of the agent's compressed history.  Fig.~\ref{fig:mnist_movie} shows the per-component movie strip: each row tracks one digit class's mean through the full protocol.  Remarkably, all three digit identities are maintained across the entire $0\to 1$ interval --- digit~0 remains recognisably~0, digit~3 remains~3, digit~8 remains~8 --- even at the oldest portion ($t\approx 0$) of the protocol.  The visual degradation is primarily in sharpness and contrast rather than class identity.

\begin{figure}[h!]
\centering
\includegraphics[width=0.55\linewidth]{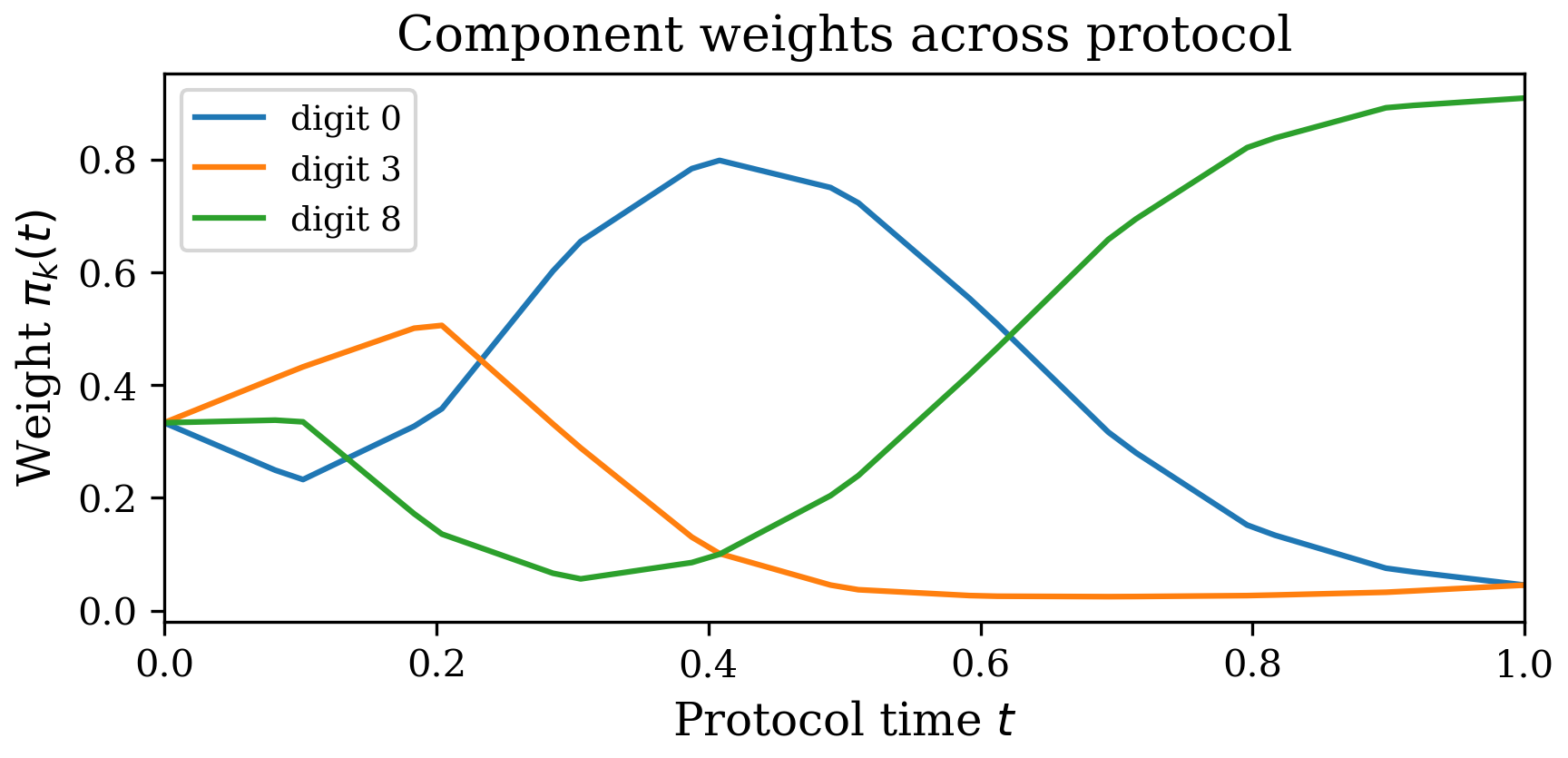}
\caption{Component weights $\pi_k(t)$ across the protocol grid.  The weight trajectory encodes a compressed history of the 100-day rotating-dominance curriculum: recent dominance of digit~8 (green at $t=1$), earlier dominance of digit~0 (blue peak at $t\approx 0.4$), and convergence toward uniform weights at the oldest end.}
\label{fig:mnist_protocol_weights}
\end{figure}

The protocol weight evolution (Fig.~\ref{fig:mnist_protocol_weights}) shows the compressed temporal history of the rotating curriculum.  At $t=1$ (current day), digit~8 dominates ($\pi_8\approx 0.9$).  Moving leftward: digit~0 peaked around $t\approx 0.4$, digit~3 around $t\approx 0.2$, and the oldest memories ($t\approx 0$) have roughly equal weights.  This weight trajectory is a lossy but recognisable compression of the full 100-day curriculum.

\subsection{Dimension sweep}

\begin{figure}[h!]
\centering
\includegraphics[width=0.75\linewidth]{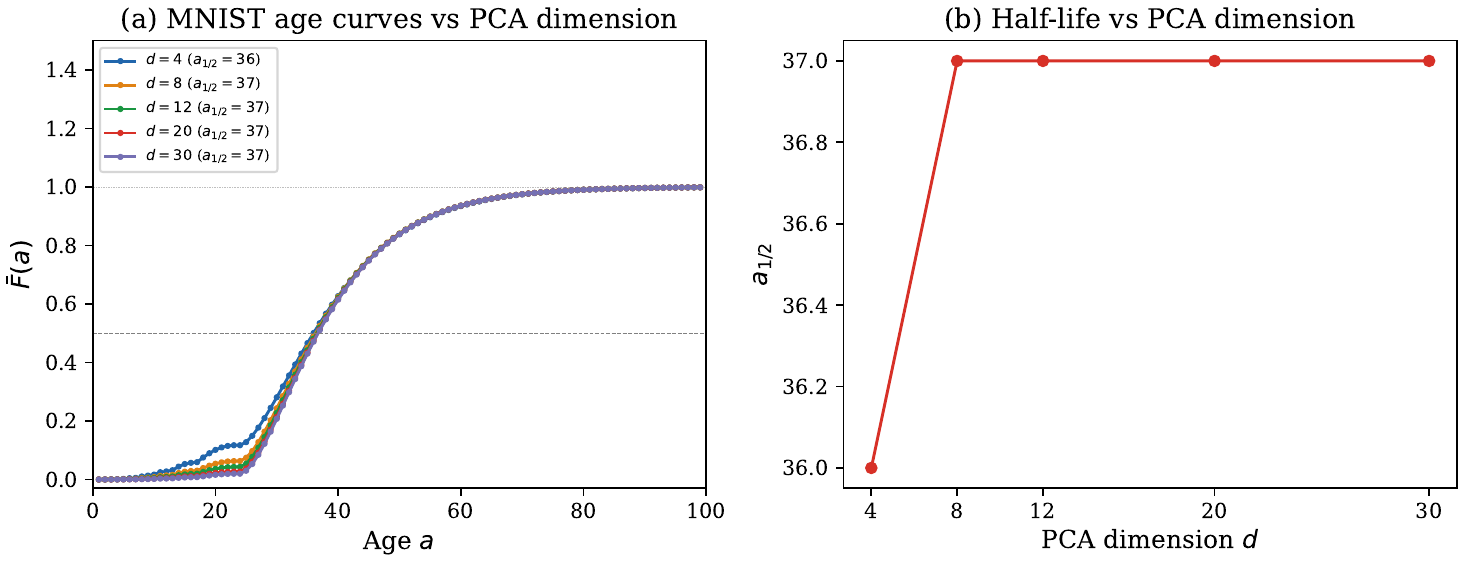}
\caption{MNIST PCA dimension sweep.  (a)~Age--forgetting curves for $d\in\{4, 8, 12, 20, 30\}$ nearly coincide.  (b)~Half-life is flat at $a_{1/2}\approx 37$ across all dimensions.}
\label{fig:mnist_dim_sweep}
\end{figure}

Sweeping the PCA dimension $d\in\{4, 8, 12, 20, 30\}$ yields half-lives $a_{1/2}\in\{36, 37, 37, 37, 37\}$ (Fig.~\ref{fig:mnist_dim_sweep}) --- essentially flat across all dimensions tested.  No semantic collapse occurs at any~$d$.  This confirms the dimension-independence observed in the synthetic scaling experiments (Section~\ref{sec:scaling}) and demonstrates that the CAS framework handles real image-derived latent spaces as robustly as synthetic ones.

\subsection{Takeaway}

The MNIST experiment demonstrates four results.  First, the CAS framework transfers successfully from synthetic to image-derived latent spaces: the two-regime forgetting curve and $a_{1/2}\approx c\,L$ scaling persist.  Second, the dominant forgetting channel shifts from mean-dominated (synthetic, where means drift) to covariance-dominated (MNIST, where means are fixed and only weights rotate) --- the framework correctly identifies the active information channel in each case.  Third, the protocol grid serves as a genuine temporal movie: decoded frame-by-frame, it produces a visual narrative in which digit identities are preserved while mixing proportions evolve smoothly from the agent's oldest memories to its most recent experience.  Fourth, there is no critical dimension for semantic collapse: the half-life is flat from $d=4$ to $d=30$, confirming that temporal compression --- not representational capacity --- is the binding constraint on retention.

\section{Discussion}
\label{sec:discussion}

We now discuss two cross-cutting themes that emerge from the full suite of experiments: the information-theoretic structure of the $a_{1/2}\approx c\,L$ law, and the role of the stochastic process underlying the bridge as a mechanism for temporally coherent replay.

\subsection{Retention capacity and the $a_{1/2}\approx c\,L$ law}
\label{sec:capacity}

The empirical law $a_{1/2}\approx 2.4\,L$, first observed in the $K=1$ experiments (Section~\ref{sec:phase1}) and confirmed unchanged across mixture complexity (Section~\ref{sec:phase2}), crowding, dimension, curriculum (Section~\ref{sec:scaling}), and MNIST (Section~\ref{sec:mnist_latent}), deserves closer examination.

\paragraph{Why $c>1$ matters.}
A First-In-First-Out (FIFO) buffer --- the simplest baseline, which stores the last $L$ daily distributions verbatim and discards the oldest upon each new arrival --- provides perfect recall for $L$~days and instant amnesia thereafter, giving $a_{1/2}=L$ exactly.  The CAS scheme achieves $a_{1/2}\approx 2.4\,L$ --- a factor $c\approx 2.4$ improvement --- despite using the same $O(LKd^2)$ storage.  The gain arises because the piecewise-linear interpolant between GM nodes implicitly encodes information about \emph{intermediate} days that do not sit on any grid node.  A readout at time $t_{m|n}$ between two nodes returns a meaningful blend of the flanking node states, carrying compressed but non-trivial information about the original day-$m$ target.  The bridge is performing lossy compression, but it is a \emph{smooth} compression whose interpolation structure extracts more than one ``effective day'' of retention per grid node.

\paragraph{Where the factor $c$ comes from.}
The origin of~$c$ can be understood from the readout-time geometry.  For large~$L$, the compression ratio per step is $L/(L+1)\approx e^{-1/L}$, so after $a$~days a memory's readout time has decayed to $t_{m|n} \approx e^{-a/L}$.  Forgetting sets in when $t_{m|n}$ drops below a critical threshold~$t_*$ at which the cumulative rebinning error crosses the forgetting criterion~$\theta=0.5$.  This gives $a_{1/2} \approx L\,\ln(1/t_*)$, identifying $c = \ln(1/t_*)$.  For $c\approx 2.4$, we get $t_*\approx 0.09$ --- consistent with the observation that a 30-day-old memory at $L=10$ sits at $t\approx 0.047$ (well past the threshold), while a 20-day-old memory sits at $t\approx 0.13$ (just above it).

The threshold~$t_*$ depends on the drift speed: faster drift raises~$t_*$ and lowers~$c$ (the speed sweep gives $c\approx 2.0$ at $P=25$ to $c\approx 3.6$ at $P=200$).  It does not depend on~$K$, $d$, or the geometry of the target family --- explaining the remarkable universality of the linear law across all our experiments.

\paragraph{Information-theoretic interpretation.}
The linear law $a_{1/2}=c\,L$ has a natural information-theoretic reading.  The protocol grid with $L$~nodes is a fixed-capacity ``channel'' of $O(LKd^2)$ real numbers; each daily incorporation injects $O(Kd^2)$ numbers.  The maximum retention is therefore $O(L)$, establishing that linear scaling is fundamental.  The constant~$c$ quantifies how efficiently the encoding utilises the available capacity --- analogous to the capacity constant in Shannon's noisy-channel coding theorem\footnote{The noisy-channel coding theorem~\cite{richardson_modern_2008} establishes that reliable communication over a noisy channel is possible at any rate below the channel capacity~$C$, but not above.  In our setting, the ``channel'' is the $L$-node protocol grid corrupted by rebinning noise, and $c$~plays the role of~$C$: it is the maximum number of effective retention days per grid node achievable by any CAS-type encoding.  The connection to rate-distortion theory~\cite{richardson_modern_2008} is even more direct: the CAS recursion trades off temporal resolution (rate) against forgetting quality (distortion) under a fixed memory budget.}.  The current uniform-grid scheme achieves $c\approx 2.4$; the question is how close an optimised scheme can get to the theoretical maximum $c_{\rm opt}$.

Three concrete optimisation avenues are:
\begin{itemize}[nosep]
\item \emph{Non-uniform (logarithmic) grids.}  Placing nodes at times $t_j = e^{-j\alpha}$ instead of $j/L$ would allocate finer resolution to recent memories and directly increase~$c$.
\item \emph{Variational rebinning.}  Optimising the node placement at each step to minimise KL divergence from the augmented protocol would define $c_{\rm opt}$ operationally.
\item \emph{Non-linear interpolation.}  Wasserstein geodesics between nodes could preserve more geometric structure through rebinning.
\end{itemize}

We conjecture that for any CAS-type scheme with $L$~grid nodes and a stationary source with characteristic drift rate~$v$, the retention half-life satisfies
\begin{equation}
  a_{1/2} \;\le\; c_{\rm opt}(v)\cdot L,
  \label{eq:capacity_bound}
\end{equation}
where $c_{\rm opt}(v)$ is a source-dependent constant.  Determining $c_{\rm opt}$ is an open problem connecting the CAS framework to rate-distortion theory.

\subsection{The role of stochastic replay}
\label{sec:stochastic_replay}

The protocol grid stores a density path $p_t(x)$ for $t\in[0,1]$.  Appendix~\ref{app:p-to-s} reconstructs a drift $s_t(x)$ such that the SDE
\begin{equation}
  dX_t = s_t(X_t)\,dt + dW_t
  \label{eq:replay_sde}
\end{equation}
has marginal density exactly~$p_t$.  This construction is never needed during the daily CAS update --- it is invoked only at ``inference time'' when sample paths are requested --- but it provides qualitative capabilities that go beyond evaluating marginal densities at readout times.

\paragraph{Replay as a movie.}
Sampling $X_0\sim p_0$ and integrating~\eqref{eq:replay_sde} forward to $t=1$ generates a continuous stochastic trajectory that visits, in temporal order, compressed representations of older memories (small~$t$), progresses through intermediate eras, and arrives at the current day ($t=1$).  Each realisation is a different plausible ``narrative'' connecting the agent's past to its present.  The MNIST experiment (Section~\ref{sec:mnist_latent}) makes this literal: the protocol grid decoded frame-by-frame produces an actual visual movie of the agent's compressed digit-class history.

\paragraph{Temporal coherence across readout times.}
Two independent evaluations of~$p_t$ at different readout times yield statistically independent marginal samples.  In contrast, a single sample path $\{X_t\}_{0\le t\le 1}$ produces \emph{correlated} samples at multiple readout times --- the replay at day~$m$ and day~$m'$ come from the same trajectory and are therefore dynamically consistent.  This temporal coherence is essential for downstream tasks that require more than pointwise recall.

\paragraph{Connection to sleep replay.}
The SDE generates compressed temporal sequences with stochastic variation --- structurally analogous to hippocampal replay during sleep, where the brain replays compressed experience sequences to consolidate memory~\cite{gonzalez_can_2020,golden_sleep_2022}.  In this analogy, the protocol grid is the memory substrate, the SDE integration is the replay episode, and the diffusion noise~$dW_t$ corresponds to the variability across replay episodes.

\paragraph{Cost separation.}
A key design feature is the separation between the cheap CAS loop ($O(LKd^2)$ per day, no sampling) and the expensive SDE integration (needed only on demand).  Evaluating $s_t(x)$ requires computing $\nabla\log p_t(x)$ and, for time-varying weights, the Poisson correction~$\nabla\psi_t(x)$ from~\eqref{eq:gmm_poisson_1d_integral} --- a cost of $O(Kd^2)$ per evaluation point per time step.  For microcontroller-class hardware, the daily update runs in real time while movie generation can be deferred to periods of low computational load --- a natural parallel to the sleep/wake dichotomy in biological memory consolidation.

\subsection{Relation to prior work}
\label{sec:related_work}

Catastrophic interference \cite{mccloskey_catastrophic_1989} -- or catastrophic forgetting \cite{french_catastrophic_1999} -- in sequentially trained networks has motivated four main CL paradigms: regularization (EWC~\cite{kirkpatrick_overcoming_2017}, SI~\cite{zenke_continual_2017}), replay (deep generative replay~\cite{shin_continual_2017}, brain-inspired replay~\cite{van_de_ven_brain-inspired_2020}), architecture expansion (progressive nets~\cite{rusu_progressive_2016}), and compression (Progress \& Compress~\cite{schwarz_progress_2018}) --- all address forgetting-by-interference in shared-parameter models.  Our framework is fundamentally different: forgetting arises from temporal coarse-graining rather than parameter overwriting, and the forgetting mechanism is localised in a single identifiable step (rebinning) rather than distributed across gradient updates.  Progress \& Compress~\cite{schwarz_progress_2018} is closest in spirit (it also separates a ``knowledge base'' from an ``active column''), but relies on neural distillation rather than analytical density operations.  Variational Continual Learning~\cite{nguyen_variational_2018} maintains a sequential posterior, formally similar to our compress--add step; however, it requires gradient-based updates and does not provide a closed-form forgetting analysis.  For surveys of the CL landscape, see~\cite{parisi_continual_2019,de_lange_continual_2022,wang_comprehensive_2024}.

Within the replay paradigm, recent work replaces the VAE/GAN generator of \cite{shin_continual_2017} with a denoising diffusion model, achieving higher-fidelity replay samples for class-incremental classification \cite{gao_ddgr_2023, jodelet_class-incremental_2023, meng_diffclass_2024}, object detection \cite{kim_sddgr_2024}, federated learning \cite{liang_diffusion-driven_2024}, industrial streaming data \cite{he_continual_2024}, and anomaly detection \cite{hu_replaycad_2025}. Our approach is structurally distinct from all of these: in diffusion-based generative replay, the diffusion model is a generator of past data samples, while the classifier (or RL agent) that actually does the learning is a separate network whose parameters are still updated by gradient descent and still subject to forgetting-by-interference. In the CAS framework, by contrast, the bridge diffusion is the memory — there is no separate generator and no gradient-based forgetting. The SDE protocol replaces the replay buffer entirely.

Our bridge diffusion is constructed by prescribing a density path and recovering the SDE drift from the Fokker--Planck equation (Appendix~\ref{app:p-to-s}).  This approach is related to Schr\"odinger bridges~\cite{leonard_survey_2013,chen_stochastic_2021,de_bortoli_diffusion_2021}, flow matching~\cite{lipman_flow_2023}, stochastic interpolants \cite{albergo_building_2023} and Path-Integral Diffusion~\cite{behjoo_harmonic_2025,chertkov_adaptive_2025,chertkov_generative_2025,chertkov_mean-field_2026}, but differs in that the density path is specified directly as a piecewise-linear interpolant, rather than optimised or learned.  

In the neuroscience literature, Bazhenov and collaborators~\cite{gonzalez_can_2020,golden_sleep_2022,tadros_sleep-like_2022,golden_interleaved_2025,vins_optimal_2025} show that sleep-like off-line replay prevents catastrophic forgetting by pushing synaptic weights toward joint solution manifolds.  Our SDE-based replay (Section~\ref{sec:stochastic_replay}) is structurally analogous, with the CAS protocol playing the role of the synaptic substrate and the SDE integration playing the role of the replay episode.

\section{Conclusions and Path Forward}\label{sec:conclusions}

We introduced the Compress--Add--Smooth (CAS) framework for continual learning, in which an agent's temporal memory is encoded as a Bridge Diffusion process on a fixed replay interval~$[0,1]$.  The framework is parameterised by two budgets: a state budget~$K$ (mixture complexity) and a temporal budget~$L$ (protocol segments).  Incorporating a new day costs $O(LKd^2)$ flops with no backpropagation, no stored data, and no neural networks.

The key experimental findings, for the Gaussian-mixture instantiation, are:

\begin{enumerate}[nosep]
\item \emph{Two-regime forgetting curve.}  The normalised forgetting $\bar F(a)$ exhibits a low-error plateau for recent memories followed by a steep sigmoid transition.  The retention half-life $a_{1/2}$ --- the age at which $\bar F$ crosses~$0.5$ --- is the natural summary statistic.

\item \emph{Linear scaling with~$L$ and the capacity constant $c$.}  The half-life scales as $a_{1/2} \approx c\,L$ with $c\approx 2.4$ for the default circular-drift geometry, from $a_{1/2} = 14$ at $L=5$ to $a_{1/2} = 74$ at $L=30$.  The fact that $c>1$ means the CAS scheme extracts more than one effective day of retention per grid node, outperforming a na\"ive FIFO buffer by a factor of~${\sim}2.4$.  We derived an analytical expression $a_{1/2}\approx L\ln(1/t_*)$ linking~$c$ to a readout-time resolution threshold (Section~\ref{sec:capacity}) and argued that~$c$ plays a role analogous to the Shannon channel capacity.

\item \emph{Independence of~$K$.}  Sweeping $K \in \{1,2,3,5,8\}$ at fixed $L=10$ yields $a_{1/2} \approx 30$ for all~$K$.  Temporal compression --- not state-space complexity --- controls the forgetting rate.

\item \emph{Drift speed matters, geometry less so.}  Faster drift (shorter period~$P$) reduces the half-life (equivalently, reduces~$c$), while the choice between circular and linear drift geometry affects the curve shape but not dramatically the timescale.

\item \emph{Confusion, not destruction.}  Old memories collapse toward recent eras ($\bar F > 1$) rather than reverting to the prior ($\bar F \to 1$).  This is visible both in the normalised metric and in the spatial displacement of replayed means toward the protocol interior.

\item \emph{Adaptive forgetting channel.}  The decomposed metric correctly identifies the active information channel: mean-dominated (${\sim}85\%$) when component means drift (synthetic experiments), covariance-dominated when only weights vary (MNIST experiment).  Weight error is negligible for equal-weight mixtures.
\end{enumerate}

The stochastic process reconstructed from the density path (Appendix~\ref{app:p-to-s}, Section~\ref{sec:stochastic_replay}) provides temporally coherent replay trajectories --- compressed ``movies'' of the agent's history --- that are structurally analogous to hippocampal sleep replay in biological memory systems.  The MNIST experiment (Section~\ref{sec:mnist_latent}) made this literal: the protocol grid, decoded frame-by-frame to pixel space, produced a visual temporal narrative in which digit identities were preserved while mixing proportions evolved smoothly from oldest to most recent memories.

\paragraph{Extensions and applications.}
Several directions are immediate.
\begin{itemize}[nosep]
\item \emph{Optimising the retention constant~$c$.}  Non-uniform (logarithmic) grids, variational rebinning, and non-linear interpolants (e.g.\ Wasserstein geodesics) could increase~$c$ beyond~2.4.  Determining the theoretical maximum $c_{\rm opt}$ connects the CAS framework to rate-distortion theory (Section~\ref{sec:capacity}).
\item \emph{Neural density families.}  The CAS recursion applies to any density class admitting interpolation.  Extending it to normalising flows or score-based models would enable high-dimensional, structured data beyond the GM class.
\item \emph{Power systems.}  A dynamic memory agent could maintain a temporally compressed history of generation/consumption probability densities over a 24-hour cycle, providing input for on-the-fly operational optimisation.
\item \emph{Lagrangian turbulence.}  Continual learning of particle-tracking statistics could carry information from small-scale to large-scale dynamics via progressively coarsened temporal representations.
\item \emph{Sleep-replay applications.}  The SDE-based replay (Section~\ref{sec:stochastic_replay}) could be used for off-line trajectory generation in model-based reinforcement learning, where temporally coherent experience replay is known to improve sample efficiency.
\end{itemize}

\section*{Acknowledgments}

The author is grateful to Maxim Bazhenov for many inspiring discussions. The author thanks the University of Arizona start-up programme for financial support. Large language models (Claude, Anthropic; ChatGPT, OpenAI) assisted with text editing and code refactoring; all mathematical derivations, scientific claims, and code were independently verified by the author.

\appendix

\section{Density Interpolants}\label{app:p-to-s}

Assume that the density path $p_t(x)$, $t\in[0,1]$, $x\in\mathbb R^d$, is known. We seek a unit-diffusion It\^o process
\begin{equation}
  dX_t = s_t(X_t)\,dt + dW_t,
  \label{eq:SDE-gmm-t}
\end{equation}
whose density is exactly $p_t(x)$. This construction — recovering an SDE drift from a prescribed density path via the Fokker–Planck equation — follows the stochastic interpolant framework of \cite{albergo_building_2023}, specialized to a piecewise-linear GM interpolant with unit diffusion coefficient.

The drift $s_t$ must satisfy the Fokker--Planck equation
\begin{equation}
    \partial_t p_t(x)+\nabla\cdot J_t(x)=0,
    \qquad
    J_t(x)\doteq s_t(x)\,p_t(x)-\frac12\nabla p_t(x),
  \label{eq:FP}
\end{equation}
where the first relation is the continuity equation and the second defines the probability current $J_t(x)$. Once the continuity equation is solved -- that is the current is expressed via the density -- a valid drift is recovered as
\begin{equation}
  s_t(x)=\frac{J_t(x)}{p_t(x)}+\frac12\nabla\log p_t(x).
  \label{eq:drift-from-current}
\end{equation}

\subsection{Densities which are Gaussian Mixtures}

Consider now the case when the density is a Gaussian mixture of degree $K$:
\begin{equation}
  p_t(x)= \sum_{k=1}^K \pi_k(t)\,g_k(x,t),
  \qquad
  g_k(x,t)\doteq \mathcal N(x; m_k(t),\Sigma_k(t)),
  \label{eq:gmm-t}
\end{equation}
where $\pi_k(t)>0$, $\sum_{k=1}^K\pi_k(t)=1$, $\Sigma_k(t)\succ0$, and $\pi_k(t)$, $m_k(t)$, $\Sigma_k(t)$ are assumed known.

\paragraph{Constant weights.}
If $\pi_k(t)\equiv \pi_k$, $k=1,\dots,K$, then the continuity equation can be integrated explicitly, resulting in the Gaussian-mixture expression
\begin{equation}
  J_t(x)=\sum_{k=1}^K \pi_k\,J_{k,t}(x),
  \qquad
  J_{k,t}(x)\doteq g_k(x,t)\left[\dot m_k(t)+ \frac12 \dot\Sigma_k(t)\Sigma_k^{-1}(t)\bigl(x-m_k(t)\bigr)\right].
  \label{eq:J-const-weights}
\end{equation}

\paragraph{Time-varying weights.}
In general, when the weights vary in time, we decompose the current into two parts:
\begin{equation}
  J_t(x)=J_t^{\mathrm{shape}}(x)+J_t^{\mathrm{wt}}(x).
  \label{eq:J-decomposition}
\end{equation}
The first term accounts for the motion and deformation of the Gaussian components:
\begin{equation}
  J_t^{\mathrm{shape}}(x)=\sum_{k=1}^K \pi_k(t)\,g_k(x,t)
  \left(\dot m_k(t)+ \frac12 \dot\Sigma_k(t)\Sigma_k^{-1}(t)\bigl(x-m_k(t)\bigr)\right).
  \label{eq:J-shape}
\end{equation}
The correction current associated with the time dependence of the weights satisfies
\begin{equation}
  \nabla\cdot J_t^{\mathrm{wt}}(x)=-\sum_{k=1}^K \dot\pi_k(t)\,g_k(x,t).
  \label{eq:J-weight-eq}
\end{equation}
Since $\sum_{k=1}^K \pi_k(t)=1$, we also have $\sum_{k=1}^K \dot\pi_k(t)=0$, and therefore the right-hand side of \eqref{eq:J-weight-eq} has zero total mass, which is the compatibility condition for a decaying solution on $\mathbb R^d$.

Looking for the correction current in gradient form,
\[
J_t^{\mathrm{wt}}(x)=-\nabla\psi_t(x),
\]
and decomposing
\[
\psi_t(x)=\sum_{k=1}^K \dot\pi_k(t)\,\psi_{k,t}(x),
\]
we obtain for each component the Poisson equation
\[
\Delta \psi_{k,t}(x)=g_k(x,t).
\]
Its solution can be written as the following one-dimensional integral:
\begin{equation}
\psi_t(x)=\frac{1}{2(2\pi)^{d/2}}\sum_{k=1}^K \dot\pi_k(t)\int_0^\infty\frac{\exp\!\left(-\frac12 (x-m_k(t))^T(\Sigma_k(t)+2sI)^{-1}(x-m_k(t))\right)}{\sqrt{\det(\Sigma_k(t)+2sI)}}\,ds.\label{eq:gmm_poisson_1d_integral}
\end{equation}

Therefore,
\begin{equation}
  J_t^{\mathrm{wt}}(x)=-\nabla\psi_t(x),
  \qquad
  J_t(x)=J_t^{\mathrm{shape}}(x)-\nabla\psi_t(x),
  \label{eq:J-final}
\end{equation}
and the resulting drift is
\begin{equation}
  s_t(x)
  =
  \frac{J_t^{\mathrm{shape}}(x)-\nabla\psi_t(x)}{p_t(x)}
  +\frac12\nabla\log p_t(x).
  \label{eq:drift-gmm-general}
\end{equation}
Equivalently, writing everything out,
\begin{equation}
  s_t(x)
  =
  \frac{
  \sum_{k=1}^K \pi_k(t)\,g_k(x,t)
  \left(\dot m_k(t)+ \frac12 \dot\Sigma_k(t)\Sigma_k^{-1}(t)\bigl(x-m_k(t)\bigr)\right)
  -\nabla\psi_t(x)
  }{
  \sum_{k=1}^K \pi_k(t)\,g_k(x,t)
  }
  +\frac12\nabla\log p_t(x).
  \label{eq:drift-gmm-general-expanded}
\end{equation}

\section{Software Design and Experimental Protocol}\label{app:design}

This appendix describes the software architecture underlying the experiments in Sections~\ref{sec:phase1}--\ref{sec:scaling} and the rationale for the experimental design choices. Code is available at \url{https://github.com/mchertkov/CAS-Bridge-Diffusion}.

\subsection{Core API: \texttt{bridge\_cas.py}}

The entire continual-learning pipeline is implemented in a single Python module, \texttt{bridge\_cas.py}, built on PyTorch to ensure full compatibility with automatic differentiation (autograd).  This enables sensitivity analysis --- e.g.\ $\partial a_{1/2}/\partial(\text{drift parameters})$ --- and GPU acceleration for scaling experiments.  The only non-differentiable component is the Hungarian matching used in the decomposed metric (Section~\ref{sec:metric_decomp}), which is a discrete assignment computed via \texttt{scipy}; it lies outside the main CAS loop and does not affect gradient flow.  The main classes are:

\begin{itemize}[nosep]
\item \texttt{GaussianMixture}: stores weights $\pi\in\R^K$, means $m\in\R^{K\times d}$, covariances $\Sigma\in\R^{K\times d\times d}$; provides methods for overall moments, density evaluation, and sampling.

\item \texttt{ProtocolGrid}: stores $L+1$ \texttt{GaussianMixture} node states at uniform times $\{0, 1/L, \ldots, 1\}$; implements the three CAS operations (compress, add, smooth) and piecewise-linear interpolation~\eqref{eq:pw_interp} for replay queries.

\item \texttt{ContinualMemory}: orchestrates the daily incorporate loop; maintains the readout-time dictionary~\eqref{eq:readout_update} and the history of original daily targets for metric computation.

\item \texttt{ForgetMetrics}: implements the raw~\eqref{eq:F_raw}, normalised~\eqref{eq:F_norm}, and decomposed~\eqref{eq:F_decomp} forgetting metrics, including Hungarian matching for $K>1$.
\end{itemize}

\subsection{Data format and storage}

The protocol state at any point in time is fully described by a list of $L+1$ Gaussian mixtures.  Each mixture is stored as three arrays $(\pi, m, \Sigma)$ of shapes $(K,)$, $(K,d)$, $(K,d,d)$.  The total storage per protocol snapshot is $(L+1)\times K\times(1 + d + d^2)$ floating-point numbers.  For diagnostic purposes, the full CAS history (all intermediate protocol states) can optionally be logged; in production, only the current protocol and readout-time dictionary are retained.

\subsection{Experimental protocol}

Each experiment follows a common workflow:
\begin{enumerate}[nosep]
\item \emph{Generate daily targets.}  A stream of $n$ daily Gaussian-mixture distributions is generated according to a specified drift model (circular, linear, random walk, or curriculum-based).

\item \emph{Run CAS loop.}  The \texttt{ContinualMemory} object is initialised with a prior $q^{(0)}$ and segment budget $L$.  Each daily target is incorporated via one compress--add--smooth cycle.  After each day, forgetting metrics are computed for all stored past days.

\item \emph{Compute diagnostics.}  The age-averaged forgetting curve $\bar F(a)$, retention half-life $a_{1/2}$, full forgetting matrix $\bar F(m,n)$, and (for $K>1$) the decomposed metric are computed and stored.

\end{enumerate}

\subsection{Design principles}

\begin{enumerate}[nosep]
\item \emph{Density-level storage.}  The protocol stores GM states (not SDE drift coefficients or Hamiltonian parameters).  This makes the representation interpretable, cheap to query, and independent of the drift-reconstruction step (Appendix~\ref{app:p-to-s}), which is only needed when sample paths are required.

\item \emph{Modular density class.}  The API is designed so that the \texttt{GaussianMixture} class can be replaced by any density family supporting: (a)~linear interpolation of parameters, (b)~moment computation, and (c)~density evaluation.  This enables future extensions to neural density estimators.

\item \emph{Stochastic generation deferred.}  Sample-path generation (via the drift from Appendix~\ref{app:p-to-s}) is not needed during the CAS recursion; it is only invoked at evaluation time for visualisation or downstream tasks.  This saves compute during the daily update loop.

\item \emph{Sweep-friendly.}  All design parameters ($L$, $K$, $d$, drift geometry, prior) are passed as constructor arguments, enabling clean parameter-sweep loops in experiment notebooks.
\end{enumerate}


\begin{thebibliography}{10}
\expandafter\ifx\csname url\endcsname\relax
  \def\url#1{\texttt{#1}}\fi
\expandafter\ifx\csname urlprefix\endcsname\relax\def\urlprefix{URL }\fi
\providecommand{\bibinfo}[2]{#2}
\providecommand{\eprint}[2][]{\url{#2}}

\bibitem{mccloskey_catastrophic_1989}
\bibinfo{author}{McCloskey, M.} \& \bibinfo{author}{Cohen, N.~J.}
\newblock \bibinfo{title}{Catastrophic interference in connectionist networks: {The} sequential learning problem}.
\newblock In \emph{\bibinfo{booktitle}{Psychology of {Learning} and {Motivation}}}, vol.~\bibinfo{volume}{24}, \bibinfo{pages}{109--165} (\bibinfo{publisher}{Academic Press}, \bibinfo{year}{1989}).

\bibitem{french_catastrophic_1999}
\bibinfo{author}{French, R.~M.}
\newblock \bibinfo{title}{Catastrophic forgetting in connectionist networks}.
\newblock \emph{\bibinfo{journal}{Trends in Cognitive Sciences}} \textbf{\bibinfo{volume}{3}}, \bibinfo{pages}{128--135} (\bibinfo{year}{1999}).

\bibitem{kirkpatrick_overcoming_2017}
\bibinfo{author}{Kirkpatrick, J.} \emph{et~al.}
\newblock \bibinfo{title}{Overcoming catastrophic forgetting in neural networks}.
\newblock \emph{\bibinfo{journal}{Proceedings of the National Academy of Sciences}} \textbf{\bibinfo{volume}{114}}, \bibinfo{pages}{3521--3526} (\bibinfo{year}{2017}).

\bibitem{shin_continual_2017}
\bibinfo{author}{Shin, H.}, \bibinfo{author}{Lee, J.~K.}, \bibinfo{author}{Kim, J.} \& \bibinfo{author}{Kim, J.}
\newblock \bibinfo{title}{Continual learning with deep generative replay}.
\newblock In \emph{\bibinfo{booktitle}{Advances in {Neural} {Information} {Processing} {Systems} ({NeurIPS})}}, vol.~\bibinfo{volume}{30} (\bibinfo{year}{2017}).

\bibitem{schwarz_progress_2018}
\bibinfo{author}{Schwarz, J.} \emph{et~al.}
\newblock \bibinfo{title}{Progress \& {Compress}: {A} scalable framework for continual learning}.
\newblock In \emph{\bibinfo{booktitle}{Proceedings of the 35th {International} {Conference} on {Machine} {Learning} ({ICML})}}, \bibinfo{pages}{4535--4544} (\bibinfo{year}{2018}).

\bibitem{gao_ddgr_2023}
\bibinfo{author}{Gao, R.} \& \bibinfo{author}{Liu, W.}
\newblock \bibinfo{title}{{DDGR}: {Continual} {Learning} with {Deep} {Diffusion}-based {Generative} {Replay}}.
\newblock In \emph{\bibinfo{booktitle}{Proceedings of the 40th {International} {Conference} on {Machine} {Learning}}}, vol. \bibinfo{volume}{202} of \emph{\bibinfo{series}{Proceedings of {Machine} {Learning} {Research}}}, \bibinfo{pages}{10744--10763} (\bibinfo{publisher}{PMLR}, \bibinfo{year}{2023}).
\newblock \urlprefix\url{https://proceedings.mlr.press/v202/gao23e.html}.

\bibitem{jodelet_class-incremental_2023}
\bibinfo{author}{Jodelet, Q.}, \bibinfo{author}{Liu, X.}, \bibinfo{author}{Phua, Y.~J.} \& \bibinfo{author}{Murata, T.}
\newblock \bibinfo{title}{Class-{Incremental} {Learning} using {Diffusion} {Model} for {Distillation} and {Replay}}.
\newblock In \emph{\bibinfo{booktitle}{Proceedings of the {IEEE}/{CVF} {International} {Conference} on {Computer} {Vision} {Workshops} ({ICCVW})}}, \bibinfo{pages}{3417--3425} (\bibinfo{publisher}{IEEE}, \bibinfo{year}{2023}).
\newblock \urlprefix\url{https://arxiv.org/abs/2306.17560}.

\bibitem{meng_diffclass_2024}
\bibinfo{author}{Meng, Z.} \emph{et~al.}
\newblock \bibinfo{title}{{DiffClass}: {Diffusion}-{Based} {Class} {Incremental} {Learning}}.
\newblock In \emph{\bibinfo{booktitle}{Computer {Vision} – {ECCV} 2024}}, vol. \bibinfo{volume}{15145} of \emph{\bibinfo{series}{Lecture {Notes} in {Computer} {Science}}}, \bibinfo{pages}{142--159} (\bibinfo{publisher}{Springer}, \bibinfo{year}{2024}).
\newblock \urlprefix\url{https://link.springer.com/chapter/10.1007/978-3-031-73021-4_9}.

\bibitem{kim_sddgr_2024}
\bibinfo{author}{Kim, J.}, \bibinfo{author}{Cho, H.}, \bibinfo{author}{Kim, J.}, \bibinfo{author}{Tiruneh, Y.~Y.} \& \bibinfo{author}{Baek, S.}
\newblock \bibinfo{title}{{SDDGR}: {Stable} {Diffusion}-based {Deep} {Generative} {Replay} for {Class} {Incremental} {Object} {Detection}}.
\newblock In \emph{\bibinfo{booktitle}{Proceedings of the {IEEE}/{CVF} {Conference} on {Computer} {Vision} and {Pattern} {Recognition} ({CVPR})}} (\bibinfo{publisher}{IEEE}, \bibinfo{year}{2024}).
\newblock \urlprefix\url{https://arxiv.org/abs/2402.17323}.

\bibitem{liang_diffusion-driven_2024}
\bibinfo{author}{Liang, J.} \emph{et~al.}
\newblock \bibinfo{title}{Diffusion-{Driven} {Data} {Replay}: {A} {Novel} {Approach} to {Combat} {Forgetting} in {Federated} {Class} {Continual} {Learning}}.
\newblock In \emph{\bibinfo{booktitle}{Computer {Vision} – {ECCV} 2024}}, Lecture {Notes} in {Computer} {Science} (\bibinfo{publisher}{Springer}, \bibinfo{year}{2024}).
\newblock \urlprefix\url{https://arxiv.org/abs/2409.01128}.

\bibitem{he_continual_2024}
\bibinfo{author}{He, J.} \emph{et~al.}
\newblock \bibinfo{title}{Continual {Learning} with {Diffusion}-based {Generative} {Replay} for {Industrial} {Streaming} {Data}}.
\newblock In \emph{\bibinfo{booktitle}{2024 {IEEE}/{CIC} {International} {Conference} on {Communications} in {China} ({ICCC})}} (\bibinfo{publisher}{IEEE}, \bibinfo{year}{2024}).
\newblock \urlprefix\url{https://arxiv.org/abs/2406.15766}.

\bibitem{hu_replaycad_2025}
\bibinfo{author}{Hu, L.} \emph{et~al.}
\newblock \bibinfo{title}{{ReplayCAD}: {Generative} {Diffusion} {Replay} for {Continual} {Anomaly} {Detection}}.
\newblock In \emph{\bibinfo{booktitle}{Proceedings of the 34th {International} {Joint} {Conference} on {Artificial} {Intelligence} ({IJCAI})}} (\bibinfo{year}{2025}).
\newblock \urlprefix\url{https://www.ijcai.org/proceedings/2025/328}.

\bibitem{parisi_continual_2019}
\bibinfo{author}{Parisi, G.~I.}, \bibinfo{author}{Kemker, R.}, \bibinfo{author}{Part, J.~L.}, \bibinfo{author}{Kanan, C.} \& \bibinfo{author}{Wermter, S.}
\newblock \bibinfo{title}{Continual lifelong learning with neural networks: {A} review}.
\newblock \emph{\bibinfo{journal}{Neural Networks}} \textbf{\bibinfo{volume}{113}}, \bibinfo{pages}{54--71} (\bibinfo{year}{2019}).

\bibitem{de_lange_continual_2022}
\bibinfo{author}{De~Lange, M.} \emph{et~al.}
\newblock \bibinfo{title}{A continual learning survey: {Defying} forgetting in classification tasks}.
\newblock \emph{\bibinfo{journal}{IEEE Transactions on Pattern Analysis and Machine Intelligence}} \textbf{\bibinfo{volume}{44}}, \bibinfo{pages}{3366--3385} (\bibinfo{year}{2022}).

\bibitem{wang_comprehensive_2024}
\bibinfo{author}{Wang, L.}, \bibinfo{author}{Zhang, X.}, \bibinfo{author}{Su, H.} \& \bibinfo{author}{Zhu, J.}
\newblock \bibinfo{title}{A comprehensive survey of continual learning: {Theory}, method and application}.
\newblock \emph{\bibinfo{journal}{IEEE Transactions on Pattern Analysis and Machine Intelligence}} \textbf{\bibinfo{volume}{46}}, \bibinfo{pages}{5362--5383} (\bibinfo{year}{2024}).

\bibitem{leonard_survey_2013}
\bibinfo{author}{Léonard, C.}
\newblock \bibinfo{title}{A survey of the {Schrödinger} problem and some of its connections with optimal transport} (\bibinfo{year}{2013}).
\newblock \urlprefix\url{http://arxiv.org/abs/1308.0215}.
\newblock \bibinfo{note}{ArXiv:1308.0215 [math]}.

\bibitem{chen_stochastic_2021}
\bibinfo{author}{Chen, Y.}, \bibinfo{author}{Georgiou, T.~T.} \& \bibinfo{author}{Pavon, M.}
\newblock \bibinfo{title}{Stochastic {Control} {Liaisons}: {Richard} {Sinkhorn} {Meets} {Gaspard} {Monge} on a {Schrödinger} {Bridge}}.
\newblock \emph{\bibinfo{journal}{SIAM Review}} \textbf{\bibinfo{volume}{63}}, \bibinfo{pages}{249--313} (\bibinfo{year}{2021}).
\newblock \urlprefix\url{https://epubs.siam.org/doi/10.1137/20M1339982}.

\bibitem{de_bortoli_diffusion_2021}
\bibinfo{author}{De~Bortoli, V.}, \bibinfo{author}{Thornton, J.}, \bibinfo{author}{Heng, J.} \& \bibinfo{author}{Doucet, A.}
\newblock \bibinfo{title}{Diffusion {Schrödinger} {Bridge} with {Applications} to {Score}-{Based} {Generative} {Modeling}}.
\newblock In \bibinfo{editor}{Ranzato, M.}, \bibinfo{editor}{Beygelzimer, A.}, \bibinfo{editor}{Dauphin, Y.}, \bibinfo{editor}{Liang, P.~S.} \& \bibinfo{editor}{Vaughan, J.~W.} (eds.) \emph{\bibinfo{booktitle}{Advances in {Neural} {Information} {Processing} {Systems}}}, vol.~\bibinfo{volume}{34}, \bibinfo{pages}{17695--17709} (\bibinfo{publisher}{Curran Associates, Inc.}, \bibinfo{year}{2021}).
\newblock \urlprefix\url{https://proceedings.neurips.cc/paper_files/paper/2021/file/940392f5f32a7ade1cc201767cf83e31-Paper.pdf}.

\bibitem{lipman_flow_2023}
\bibinfo{author}{Lipman, Y.}, \bibinfo{author}{Chen, R. T.~Q.}, \bibinfo{author}{Ben-Hamu, H.}, \bibinfo{author}{Nickel, M.} \& \bibinfo{author}{Le, M.}
\newblock \bibinfo{title}{Flow {Matching} for {Generative} {Modeling}} (\bibinfo{year}{2023}).
\newblock \urlprefix\url{http://arxiv.org/abs/2210.02747}.
\newblock \bibinfo{note}{ArXiv:2210.02747 [cs, stat]}.

\bibitem{albergo_building_2023}
\bibinfo{author}{Albergo, M.~S.} \& \bibinfo{author}{Vanden-Eijnden, E.}
\newblock \bibinfo{title}{Building {Normalizing} {Flows} with {Stochastic} {Interpolants}} (\bibinfo{year}{2023}).
\newblock \urlprefix\url{http://arxiv.org/abs/2209.15571}.
\newblock \bibinfo{note}{ArXiv:2209.15571 [cs, stat]}.

\bibitem{behjoo_harmonic_2025}
\bibinfo{author}{Behjoo, H.} \& \bibinfo{author}{Chertkov, M.}
\newblock \bibinfo{title}{Harmonic {Path} {Integral} {Diffusion}}.
\newblock \emph{\bibinfo{journal}{IEEE Access}} \textbf{\bibinfo{volume}{13}}, \bibinfo{pages}{42196--42213} (\bibinfo{year}{2025}).
\newblock \urlprefix\url{https://ieeexplore.ieee.org/document/10910146/}.

\bibitem{chertkov_adaptive_2025}
\bibinfo{author}{Chertkov, M.} \& \bibinfo{author}{Behjoo, H.}
\newblock \bibinfo{title}{Adaptive {Path} {Integral} {Diffusion}: {AdaPID}} (\bibinfo{year}{2025}).
\newblock \urlprefix\url{http://arxiv.org/abs/2512.11858}.
\newblock \bibinfo{note}{ArXiv:2512.11858 [cs]}.

\bibitem{chertkov_generative_2025}
\bibinfo{author}{Chertkov, M.}
\newblock \bibinfo{title}{Generative {Stochastic} {Optimal} {Transport}: {Guided} {Harmonic} {Path}-{Integral} {Diffusion}} (\bibinfo{year}{2025}).
\newblock \urlprefix\url{http://arxiv.org/abs/2512.11859}.
\newblock \bibinfo{note}{ArXiv:2512.11859 [cs]}.

\bibitem{chertkov_mean-field_2026}
\bibinfo{author}{Chertkov, M.}
\newblock \bibinfo{title}{Mean-field path-integral diffusion: {From} samples to interacting agents} (\bibinfo{year}{2026}).
\newblock \urlprefix\url{https://github.com/mchertkov/MeanFieldPID}.

\bibitem{gonzalez_can_2020}
\bibinfo{author}{González, O.~C.}, \bibinfo{author}{Sokolov, Y.}, \bibinfo{author}{Krishnan, G.~P.}, \bibinfo{author}{Delanois, J.~E.} \& \bibinfo{author}{Bazhenov, M.}
\newblock \bibinfo{title}{Can sleep protect memories from catastrophic forgetting?}
\newblock \emph{\bibinfo{journal}{eLife}} \textbf{\bibinfo{volume}{9}}, \bibinfo{pages}{e51005} (\bibinfo{year}{2020}).

\bibitem{golden_sleep_2022}
\bibinfo{author}{Golden, R.}, \bibinfo{author}{Delanois, J.~E.}, \bibinfo{author}{Sanda, P.} \& \bibinfo{author}{Bazhenov, M.}
\newblock \bibinfo{title}{Sleep prevents catastrophic forgetting in spiking neural networks by forming a joint synaptic weight representation}.
\newblock \emph{\bibinfo{journal}{PLOS Computational Biology}} \textbf{\bibinfo{volume}{18}}, \bibinfo{pages}{e1010628} (\bibinfo{year}{2022}).

\bibitem{tadros_sleep-like_2022}
\bibinfo{author}{Tadros, T.}, \bibinfo{author}{Krishnan, G.~P.}, \bibinfo{author}{Ramyaa, R.} \& \bibinfo{author}{Bazhenov, M.}
\newblock \bibinfo{title}{Sleep-like unsupervised replay reduces catastrophic forgetting in artificial neural networks}.
\newblock \emph{\bibinfo{journal}{Nature Communications}} \textbf{\bibinfo{volume}{13}}, \bibinfo{pages}{7742} (\bibinfo{year}{2022}).

\bibitem{golden_interleaved_2025}
\bibinfo{author}{Golden, R.} \emph{et~al.}
\newblock \bibinfo{title}{Interleaved replay of novel and familiar memory traces during slow-wave sleep prevents catastrophic forgetting} (\bibinfo{year}{2025}).
\newblock \bibinfo{note}{Published: bioRxiv 2025.06.25.661579}.

\bibitem{vins_optimal_2025}
\bibinfo{author}{Vins, D.}, \bibinfo{author}{Delanois, J.~E.} \& \bibinfo{author}{Bazhenov, M.}
\newblock \bibinfo{title}{Optimal stopping for continual learning}.
\newblock In \emph{\bibinfo{booktitle}{Proceedings of the {AAAI} {Conference} on {Artificial} {Intelligence}}} (\bibinfo{year}{2025}).

\bibitem{lecun_mnist_1998}
\bibinfo{author}{LeCun, Y.}, \bibinfo{author}{Bottou, L.}, \bibinfo{author}{Bengio, Y.} \& \bibinfo{author}{Haffner, P.}
\newblock \bibinfo{title}{Gradient-based learning applied to document recognition}.
\newblock \emph{\bibinfo{journal}{Proceedings of the IEEE}} \textbf{\bibinfo{volume}{86}}, \bibinfo{pages}{2278--2324} (\bibinfo{year}{1998}).

\bibitem{richardson_modern_2008}
\bibinfo{author}{Richardson, T.} \& \bibinfo{author}{Urbanke, R.}
\newblock \emph{\bibinfo{title}{Modern Coding Theory}} (\bibinfo{publisher}{Cambridge University Press}, \bibinfo{address}{Cambridge}, \bibinfo{year}{2008}).

\bibitem{zenke_continual_2017}
\bibinfo{author}{Zenke, F.}, \bibinfo{author}{Poole, B.} \& \bibinfo{author}{Ganguli, S.}
\newblock \bibinfo{title}{Continual learning through synaptic intelligence}.
\newblock In \emph{\bibinfo{booktitle}{Proceedings of the 34th {International} {Conference} on {Machine} {Learning} ({ICML})}}, \bibinfo{pages}{3987--3995} (\bibinfo{year}{2017}).

\bibitem{van_de_ven_brain-inspired_2020}
\bibinfo{author}{van~de Ven, G.~M.}, \bibinfo{author}{Siegelmann, H.~T.} \& \bibinfo{author}{Tolias, A.~S.}
\newblock \bibinfo{title}{Brain-inspired replay for continual learning with artificial neural networks}.
\newblock \emph{\bibinfo{journal}{Nature Communications}} \textbf{\bibinfo{volume}{11}}, \bibinfo{pages}{4069} (\bibinfo{year}{2020}).

\bibitem{rusu_progressive_2016}
\bibinfo{author}{Rusu, A.~A.} \emph{et~al.}
\newblock \bibinfo{title}{Progressive neural networks} (\bibinfo{year}{2016}).
\newblock \bibinfo{note}{Published: arXiv:1606.04671}.

\bibitem{nguyen_variational_2018}
\bibinfo{author}{Nguyen, C.~V.}, \bibinfo{author}{Li, Y.}, \bibinfo{author}{Bui, T.~D.} \& \bibinfo{author}{Turner, R.~E.}
\newblock \bibinfo{title}{Variational continual learning}.
\newblock In \emph{\bibinfo{booktitle}{International {Conference} on {Learning} {Representations} ({ICLR})}} (\bibinfo{year}{2018}).

\end{thebibliography}

\end{document}